\documentclass[preprint,11pt,5p,twocolumn,authoryear,]{elsarticle}

\usepackage{amssymb}
\usepackage{amsmath}

\usepackage[switch]{lineno}

\usepackage{hyperref}
\usepackage{url}
\usepackage{xcolor}
\usepackage{blindtext}
\usepackage{circuitikz}
\usepackage{tikz}
\usepackage{multirow}
\usepackage{makecell}
\usepackage{booktabs}

\begin{document}

\begin{frontmatter}



\title{Artificial Intelligence for Climate Adaptation: Using Reinforcement Learning for Climate Change-Resilient Transport} 


 \author[dtu]{Miguel Costa\corref{corresponding}} 
 \ead{migcos@dtu.dk}
 \cortext[corresponding]{Corresponding author.}
 
 \author[dtu]{Arthur Vandervoort}
 \ead{apiva@dtu.dk}
 
 \author[dtu]{Carolin Schmidt\fnref{tum}}
 \ead{carolin.schmidt@tum.de}
 
 \author[dtu]{João Miranda}
 \ead{s232660@student.dtu.dk}
 
 \author[dtu]{Morten W. Petersen\fnref{sgd}}
 \ead{mwipe@dtu.dk}
 
 \author[dtu]{Martin Drews}
 \ead{mard@dtu.dk}
 
 \author[ugalway]{Karyn Morrisey}
 \ead{karyn.morrissey@universityofgalway.ie}
 
 \author[dtu]{Francisco C. Pereira}
 \ead{camara@dtu.dk}
 
 \affiliation[dtu]{organization={Department of Technology, Management and Economics, Technical University of Denmark},
             city={2800 Kgs. Lyngby},
             country={Denmark}}
 \affiliation[ugalway]{organization={J.E. Cairnes School of Business and Economics, University of Galway},
             city={Galway},
             postcode={H91 TK33},
             country={Ireland}}
 \fntext[tum]{\textit{Present address}: School of Management, Technical University of Munich, 80333 Munich, Germany.}
 \fntext[sgd]{\textit{Present address}: Saint-Gobain Distribution Denmark, 2605 Brøndby, Denmark.}

\begin{abstract}
Climate change is expected to intensify rainfall and, consequently, pluvial flooding, leading to increased disruptions in urban transportation systems over the coming decades. Designing effective adaptation strategies is challenging due to the long-term, sequential nature of infrastructure investments, deep climate uncertainty, and the complex interactions between flooding, infrastructure, and mobility impacts. In this work, we propose a novel decision-support framework using reinforcement learning (RL) for long-term flood adaptation planning. Formulated as an integrated assessment model (IAM), the framework combines rainfall projection and flood modeling, transport simulation, and quantification of direct and indirect impacts on infrastructure and mobility. Our RL-based approach learns adaptive strategies that balance investment and maintenance costs against avoided impacts. We evaluate the framework through a case study of Copenhagen's inner city over the 2024--2100 period, testing multiple adaptation options, and different belief and realized climate scenarios. Results show that the framework outperforms traditional optimization approaches by discovering coordinated spatial and temporal adaptation pathways and learning trade-offs between impact reduction and adaptation investment, yielding more resilient strategies. Overall, our results showcase the potential of reinforcement learning as a flexible decision-support tool for adaptive infrastructure planning under climate uncertainty.
\end{abstract}


\begin{keyword}
Climate Adaptation \sep 
Reinforcement Learning \sep 
Pluvial Flooding \sep 
Transportation \sep
Climate Change \sep 
Copenhagen (Denmark)
\end{keyword}

\end{frontmatter}


\section{Introduction}
\label{sec:introduction}

From 1998 to 2017, floods have accounted for 43\% of all recorded climate-related and geophysical disasters, affecting about 2 billion people \citep{wallemacq2018economic}. As climate change and global warming intensify, the frequency of heavy precipitation and flooding events is expected to increase \citep{ipcc2023climate}. In Denmark, such impacts are also expected, with extreme precipitation becoming more extreme and frequent \citep{dmi2011adaptation}. Reducing the risk posed by such events is vital to protecting social and economic activities. For many cities worldwide, this has become a crucial task of governing and managing an economically viable and liveable city \citep{pregnolato2017impact}. 

Urban transportation systems and networks are particularly vulnerable to floods due to their increased concentration and vulnerability to weather-related hazards \citep{yin2016evaluating, pregnolato2017climate, singh2018vulnerability, he2021flood}. Flooding can directly and indirectly impact transportation systems, causing, e.g., delays, loss of vehicle control, re-routing, or even inducing road closures \citep{wang2020climate}. In 2011, Copenhagen, Denmark (our case study), saw a cloudburst event flood many of the city's streets, leading to road and highway closures, train constraints, and cancellations \citep{hvass2011sadan}, causing an estimated 6 billion Danish kroner, DKK (805 million Euros) damage cost \citep{gerdes2012what}. Thus, it is vital to reduce flood-related risks. Investments and interventions should capture mitigation and adaptation value so that long-term benefits can be integrated into cost-benefit analysis and to weigh the increased resilience of both transport and flood risk management infrastructures \citep{pregnolato2016assessing}. Understanding the impacts and costs of such hazards is critical to enhancing resilience and ensuring the safety of a city's population. 

Overall, it is vital that not only are impacts correctly estimated and measured, but that we can identify interventions that adapt our cities to minimise such flood-related impacts. However, while research has examined how to devise short-term strategies to increase cities' resilience to floods, long-term (i.e., 50-100-year time horizon) planning has remained relatively unexplored. This limits the effectiveness of adaptation strategies, given current and potential future climate projections. 
In this work, we address this gap by proposing a reinforcement learning–based AI framework. In effect, we build a framework that incorporates current rainfall and flood projections, models how trips are disrupted by such events, and computes direct and indirect impacts on transport. To our knowledge, this is the first comprehensive framework for identifying the best adaptation measures to enhance long-term transportation resilience to urban floods. 
This work's main objectives are threefold:
\begin{itemize}
    \item We build a framework that integrates a rainfall projection model, a flood model, a transport simulation model, and a component that measures direct and indirect impacts of floods on transport;

    \item Use reinforcement learning to learn what the best long-term transport-related adaptation measures are that minimise pluvial floods' impacts on transport; and

    \item Test our framework under different climate projection scenarios and assess different adaptation pathways.
\end{itemize}

The article is structured as follows. After this introductory section, we provide some background on related work in Section \ref{sec:background}. Methodology is then explained in Section \ref{sec:methods}. We compare our approach versus state of the art benchmarks in Section \ref{sec:methods_validation}. Next, we present our main results in Section \ref{sec:results}, which are then discussed in Section \ref{sec:discussion}. Section \ref{sec:conclusions} presents overall conclusions and some leads for possible future research.

\section{Background}
\label{sec:background}

\subsection{Floods and transport}

Floods are often divided by their cause (e.g., precipitation, water damming, or failures of technical infrastructure) \citep{smith1998floods}, impact severity (e.g., different severity classes or scales) \citep{diakakis2020proposal}, or types (e.g., river flood, flash flood, coastal flood) \citep{georgi2012urban}. The particular case of flash floods in urban areas occurs when a large volume of water accumulates in streets or roads due to limited soil water absorption or insufficient capacity of drainage systems and water storage \citep{borowska2024changes}. These events can be caused by heavy precipitation or cloudburst events. Cloudbursts are defined by the Danmarks Meteorologiske Institut (Danish Meteorological Institute) as more than 15 mm of rainfall in under 30 minutes \citep{dmi2021van}, causing impacts to society, both in terms of losses of lives and physical damage \citep{schmith2023regional}.

Flooding impacts on transportation can be divided into four categories: direct physical impacts (e.g., road deterioration or washout), direct non-physical impacts (e.g., transport delay and congestion), indirect physical impacts (e.g., public transportation interruption due to electric power shortages), and indirect non-physical impacts (e.g., accessibility loss or disruption caused by failure of information communication technologies) \citep{markolf2019transportation, lu2022overview}. Understanding the challenges posed by floods and their ``paths of disruption'' to transportation, both direct and indirect, may serve as the first step toward defining robust, resilient, flexible, and agile solutions and helping decision-makers choose the most suitable plan of action to minimise such disruptions \citep{markolf2019transportation}. 

To this end, several studies have explored direct and indirect impacts on transportation, with most flood-related economic studies analysing direct costs of restoring or rebuilding infrastructure or public agencies' costs \citep{chang2011future}, or indirect costs of travel delays and congestion \citep{shahdani2022assessing, ding_interregional_2023}. These types of costs should be considered when conducting a cost-benefit analysis (CBA) to compare different solutions or infrastructure/project investments, such as flood-prevention interventions. Planners often prioritize short-term decisions with more limited consequences, and may rely on public-facing agreements or non-transparent analytical processes \citep{Hutter01082007}. In turn, this limits comprehensive analyses of flooding and possible interventions to specific flood scenarios or to the implementation of long-term (50--100 years) action plans. Altogether, this means that current and future projections of climate events are often not considered in flood risk management, thereby limiting the effectiveness of interventions when a longer time horizon is considered.

\subsection{Artificial intelligence for climate adaptation}

Climate adaptation refers to "the process of adjustment to actual or expected \textit{climate} and its effects, in order to moderate harm or exploit beneficial opportunities" \citep{intergovernmentalpanelonclimatechangeipccClimateChange20222023a}. Adaptation has the potential to reduce immediate and future vulnerabilities to both expected average changes and possible extreme events \citep{smit2001adaptation}. However, climate adaptation planning and development are not simple, requiring models that explicitly account for trade-offs inherent to different adaptation strategies \citep{garner_climate_2016}. In this role, artificial intelligence (AI) tools can address this challenge by serving as decision-support frameworks that integrate complex dynamics, synergies, and trade-offs, heterogeneous actors, and unknown impacts \citep{cheong2022artificial}, allowing for better integration of scientific evidence into policymaking \citep{tyler2023ai}.

AI and machine learning (ML) can be used to address many climate change-related areas, including transportation, industry, carbon dioxide capture and removal, electricity, or farms and forest management, by leveraging advances in different fields and methods from causal inference, computer vision, reinforcement learning, time-series analysis, or uncertainty quantification \citep{rolnick2022tackling}. AI can also help in data-scarce problems pervasive to adaptation research \citep{cheong2022artificial}. Climate change impacts take time to emerge, but impact assessment needs to happen in advance to protect social and physical systems from potential harm. AI frameworks have shown promise for limited-data tasks using transfer learning, semi-supervised learning, multiarmed bandits, or reinforcement learning \citep{cheong2022artificial}.

Reinforcement learning, in particular, can help navigate complex dynamics by identifying optimal policies that balance short- and long-term objectives and the implications of different policy priorities \citep{gilbert2022choicesrisksrewardreports}. RL achieves this by overcoming traditional methods for handling systems' non-linearity by learning from an interactive environment and its constraints to maximise a delayed reward \citep{matsuo2022deep}. When compared to other dynamic adaptation frameworks (e.g., based on Dynamic Programming \citep{LICKLEY201418}, Bayesian Optimisation \citep{sobhaniyeh2021robust}, or Direct policy Search \citep{GARNER201896}), RL allows for developing dynamic policies that systemically considers observational data and future observations to make strategic decisions and adjust to current decision-making goals \citep{fend2025reinforcement}. 

Related to flood management and transportation problems, RL has been applied to various tasks, including emergency routing systems \citep{li2024reinforcement}, drainage and stormwater system control \citep{tian2023flooding, bowes2021flood}, and response strategy studies \citep{fan2021evaluating}. However, all these have focused on reactive strategies (responding to floods as they occur) rather than proactively analysing adaptation pathways that minimise future flooding impacts. To our knowledge, only one work has explored how RL could be used for long-term flood management. \citet{fend2025reinforcement} has employed RL for coastal flood risk management for 2000--2100. However, two key limitations exist. First, decisions (adaptation measures) can be taken only every 10 years, limiting adaptation flexibility and the exploration of possible adaptation strategies. Second, the proposed framework faces substantial computational scalability challenges due to the neural network architecture underlying the RL agent. This led the authors to focus on a very restrictive action space (e.g., single-variable seawall height). Therefore, it struggles with high-frequency, spatially dense interventions required for pluvial flood management in complex urban networks.

Overall, there is currently a gap in which exploiting RL's flexibility to uncover long-term adaptation policy pathways can yield significant value. When combined with other systems (e.g., transportation and urban planning), RL frameworks can enable more responsive and effective adaptation strategies that account for evolving climate risks, socioeconomic changes, and infrastructure development. This approach would allow policymakers to explore a broader range of intervention combinations, both in scale and variety, and ultimately identify more robust and cost-effective pathways for climate resilience.

\section{Problem Formulation}
\label{sec:methods}

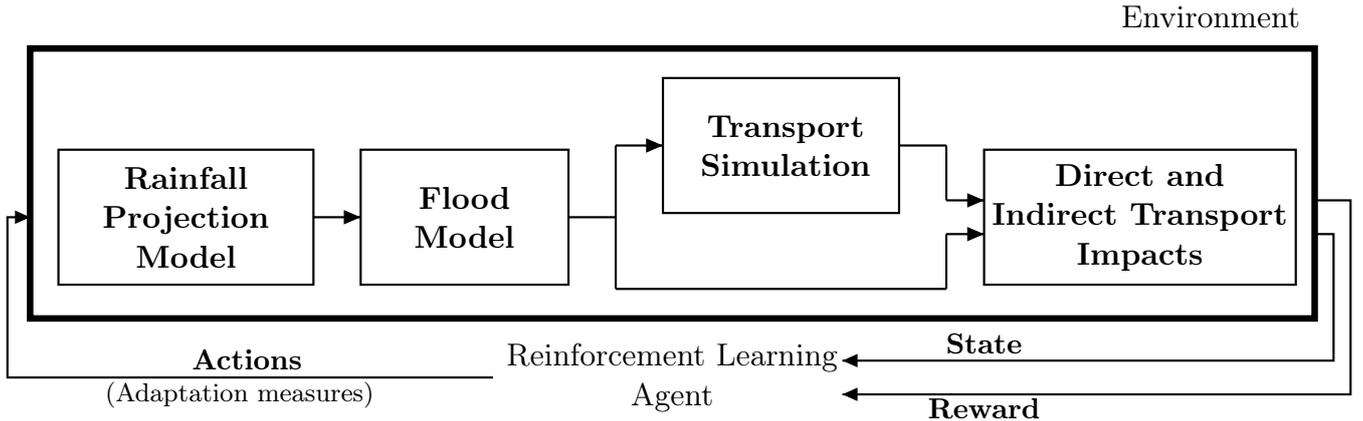
\begin{figure*}[htb]
    \centering
    \resizebox{0.98\linewidth}{!}{%
        \tikzset{every picture/.style={line width=0.75pt}} 
        \begin{tikzpicture}[x=0.65pt,y=0.65pt,yscale=-.9,xscale=1]

\draw [line width=2.25] (25,60) -- (705,60) -- (705,220) -- (25,220) -- cycle  ;

\draw (365,255) node   [align=left] {\begin{minipage}[lt]{120pt}\setlength\topsep{0pt}
\begin{center}
Reinforcement Learning Agent
\end{center}
\end{minipage}};

\draw (705,170) -- (715,170) -- (715,245) -- (455,245) ;
\draw [shift={(455,245)}, rotate = 360] [fill={rgb, 255:red, 0; green, 0; blue, 0 }  ][line width=0.08]  [draw opacity=0] (8,-4) -- (0,0) -- (8,4) -- cycle ;
\draw (705,150) -- (724,150) -- (724,265) -- (455,265) ;
\draw [shift={(455,265)}, rotate = 360] [fill={rgb, 255:red, 0; green, 0; blue, 0 }  ][line width=0.08]  [draw opacity=0] (8,-4) -- (0,0) -- (8,4) -- cycle ;
\draw (270,255) -- (13,255) -- (13,160) -- (20,160) ;
\draw [shift={(25,160)}, rotate = 180] [fill={rgb, 255:red, 0; green, 0; blue, 0 }  ][line width=0.08]  [draw opacity=0] (8,-4) -- (0,0) -- (8,4) -- cycle ;

\draw (530,235) node [font=\small] [align=left] {\textbf{State}};
\draw (530,273) node [font=\small] [align=left] {\textbf{Reward}};
\draw (136,240) node [font=\small] [align=left] {\begin{minipage}[lt]{32.83pt}\setlength\topsep{0pt}
\begin{center}
\textbf{Actions}
\end{center}
\end{minipage}};
\draw (136,265) node [font=\small] [align=left] {\begin{minipage}[lt]{100.pt}\setlength\topsep{0pt}
\begin{center}
{\footnotesize (Adaptation measures)}
\end{center}

\end{minipage}};
\draw (650,41) node [align=left] {Environment};

\draw    (40,120) -- (175,120) -- (175,200) -- (40,200) -- cycle  ;
\draw (107.5,160) node   [align=left] {\begin{minipage}[lt]{80pt}\setlength\topsep{0pt}
\begin{center}
\textbf{Rainfall Projection Model}\\{}
\end{center}
\end{minipage}};

\draw    (200,120) -- (310,120) -- (310,200) -- (200,200) -- cycle  ;
\draw (255,160) node   [align=left] {\begin{minipage}[lt]{60pt}\setlength\topsep{0pt}
\begin{center}
\textbf{Flood Model}\\
{}
\end{center}
\end{minipage}};

\draw    (360,77.5) -- (485,77.5) -- (485,157.5) -- (360,157.5) -- cycle  ;
\draw (425,117.5) node   [align=left] {\begin{minipage}[lt]{80pt}\setlength\topsep{0pt}
\begin{center}
\textbf{Transport Simulation}\\
{}
\end{center}
\end{minipage}};

\draw    (530,120) -- (695,120) -- (695,200) -- (530,200) -- cycle  ;
\draw (612.5,160) node   [align=left] {\begin{minipage}[lt]{103pt}\setlength\topsep{0pt}
\begin{center}
\textbf{Direct and Indirect Transport Impacts}\\
{}
\end{center}
\end{minipage}};

\draw    (175,160) -- (200,160) ;
\draw [shift={(200,160)}, rotate = 180] [fill={rgb, 255:red, 0; green, 0; blue, 0 }  ][line width=0.08]  [draw opacity=0] (8.93,-4.29) -- (0,0) -- (8.93,4.29) -- cycle ;
\draw    (335,117.5) -- (360,117.5) ;
\draw [shift={(360,117.5)}, rotate = 180] [fill={rgb, 255:red, 0; green, 0; blue, 0 }  ][line width=0.08]  [draw opacity=0] (8.93,-4.29) -- (0,0) -- (8.93,4.29) -- cycle ;

\draw  (485,117.5) -- (510,117.5) ;
\draw  (510,117.5) -- (510,150) ;
\draw  (510,150) -- (530,150) ;
\draw [shift={(530,150)}, rotate = 180] [fill={rgb, 255:red, 0; green, 0; blue, 0 }  ][line width=0.08]  [draw opacity=0] (8.93,-4.29) -- (0,0) -- (8.93,4.29) -- cycle ;

\draw   (335,202.5) -- (510,202.5) ;
\draw   (510,170) -- (510,202.5) ;
\draw  (510,170) -- (530,170) ;
\draw [shift={(530,170)}, rotate = 180] [fill={rgb, 255:red, 0; green, 0; blue, 0 }  ][line width=0.08]  [draw opacity=0] (8.93,-4.29) -- (0,0) -- (8.93,4.29) -- cycle ;
§
\draw   (335,117.5) -- (335,202.5) ;

\draw    (310,160) -- (335,160) ;
            
        \end{tikzpicture}
    }
    \caption{Integrated Assessment Model using reinforcement learning to find the best sequence of transport-related adaptation policies for rainfall events in Copenhagen from 2024--2100.}
    \label{fig:met_iam}
\end{figure*}

We aim to learn an adaptive policy that minimises the impact of flooding on transport using reinforcement learning. Based on the learnt policy, a sequence or portfolio of adaptation measures can be created that seeks to minimize impacts while considering expected future rainfall scenarios. To model adaptation measures and impacts, we frame our environment as an Integrated Assessment Model (IAM), as shown in Figure~\ref{fig:met_iam}. Our IAM is composed of four modules: a rainfall projection model, a flood model, a transport simulation model, and an impacts computation module. We now detail each component.

\subsection{Rainfall projection model}

The first component of our IAM is a rainfall projection model that simulates rainfall events for the 2024--2100 period. We retrieved future daily rainfall statistics under three climate scenarios (RCP2.6, conservative; RCP4.5, intermediate; RCP8.5, extreme) \citep{vanvuuren2011} from the Danish Meteorological Institute's Climate Atlas \citep{dmi2023klimaatlas} for the periods 2011--2040, 2041--2070, and 2071--2100. For each time slice (one year), we assumed stationarity and constructed the associated cumulative distribution function (CDF). Based on these CDFs, we sampled one rainfall event per timestep, represented as the accumulated daily rainfall total. This approach, combined with the climate scenario assumptions, likely overestimates rainfall intensities and therefore represents a worst-case scenario for each period. The model thus generates one rainfall event per timestep, enabling spatial mapping of the associated flooding in the urban area.

\subsection{Flood model}

For each rainfall event, we modeled the associated flooding in Copenhagen. Depending on rainfall intensity, floods can range from minor, recurring events (e.g., nuisance floods) to high-impact, major flooding events (e.g., those resulting from cloudbursts). To model all flood types across varying rainfall intensities, we used SCALGO Live~\citep{scalgo}.

SCALGO Live is a simplified, interactive, event-based tool for watershed delineation and flood depth modeling. It uses high-resolution digital terrain data, including terrain elevation \citep{danishelevationmodel} and surface properties, to map water flow direction and identify locations where water accumulates in Copenhagen. For each precipitation event, the tool assumes a uniform rainfall distribution of unspecified duration over the study area, meaning water accumulates simultaneously at all locations. Water is then redistributed according to terrain characteristics, flowing downstream and filling depressions found. When the water volume exceeds the capacity of a depression, it overflows and continues downstream. After identifying accumulation locations, we filtered water depths corresponding to roads, cycling lanes, and sidewalks (cf. Section~\ref{sec:met_transport_simulation}), enabling us to assign water heights to specific transport network locations.

\subsection{Transport simulation}
\label{sec:met_transport_simulation}

We extracted the drivable, cyclable, and walkable transport networks covering Copenhagen's inner city (\textit{indre by}) using \textsc{osmnx} \citep{boeing2024modeling} from OpenStreetMap\footnote{Available at: \url{https://www.openstreetmap.org/}} (OSM, as of July 7, 2024). OSM data has been extensively validated regarding its maturity and completeness, with studies confirming that road networks in urban areas are adequately represented \citep{barrington2017world}. The retrieved data contains georeferenced information on physical roads, cycling lanes, sidewalks, and segment-specific attributes (e.g., speed limit, number of lanes, surface type). Different transport modes have distinct infrastructure requirements. For example, pedestrians cannot use motorways so that different routes can be estimated based on each mode's constraints.

Next, to capture the impacts of flooding on transport, we modeled trips made by Copenhagen residents. We retrieved data on trips made by car, bicycle, or on foot from the Danish National Travel Survey \citep[Transportvaneundersøgelsen, TU]{christiansen2021tu}, selecting those that both originated and terminated within Copenhagen's inner city. Our study area is divided into 29 traffic assignment zones (TAZs, each covering a few city blocks), corresponding to those used in the Danish National Transport Model \citep[Grøn Mobilitetsmodel, GMM]{vejdirektoratet2022gmm}. For each trip, the origin and destination TAZ are known, but no further geographical details are available to preserve respondents' anonymity. To model each trip's route more precisely, we randomly sampled an origin and destination location within the respective TAZs. In total, our case study includes 84,000 trips: 2.7\% by car, 19.3\% by bicycle, and 78.0\% on foot.

Finally, we simulated the route each individual took, using their assigned transport mode from origin to destination on the corresponding transport network. We computed the shortest path by travel time for each trip's origin-destination pair. To account for the effect of water levels on travel, we applied depth-disruption functions that map water depths on each segment (road, street, cycling lane, sidewalk) to reduce travel speeds. We used functions introduced by \citet{pregnolato2017impact} to model speed reductions for driving and by \citet{finnis2008field} for walking. To the best of our knowledge, such functions do not exist for cycling, and, therefore, we defined a custom function for cycling between driving and walking. Each function maps the decrease in travel speed from its maximum (road speed limits for cars, 16.2km/h for bicycles \citep{bicycleaccount2022}, and 5.65km/h for walking \citep{finnis2008field}, corresponding to 0cm water depth) to its minimum (0km/h for all modes, corresponding to 30cm for cars, 20cm for bicycles, and 150~cm for walking), representing segments that are no longer traversable. This approach yields the estimated travel time \(t_j\) for each trip \(j\) under the given rainfall event.

\subsection{Transportation impacts}
Finally, after each rainfall event and its consequent flooding and transport simulation, we computed direct and indirect transportation impacts. Three types of impacts are estimated: 1) direct network infrastructure damage, 2) indirect travel delays, and 3) indirect trip cancellations.

\subsubsection{Infrastructure damage impacts}
The first type of impact we account for pertains to direct infrastructure damage caused by floods. When a flood occurs, infrastructure is damaged and should be repaired to its original condition. To account for the economic costs of these repairs, we began by estimating the overall total construction costs of each element in our network. Using the previously downloaded transport network data, we estimated construction costs for each road based on its characteristics. For this, we used an approach previously introduced by \citet{ginkel2021flood}, which estimates roads' construction costs based on road typology, number of lanes, presence of light posts and traffic lights, and then inflation- and GDP-adjusted. Next, we computed the estimated damage in each segment using depth-damage functions \citep{ginkel2021flood}. Such functions map the percentage of damage on a road depending on the water depth at each location. Such damage necessitates reconstruction, repair, cleaning, and resurfacing work to rehabilitate roads. Finally, we aggregated these costs as \(I_{i}\) monetary losses for the \(i\)-th TAZ.

\subsubsection{Travel delay impacts}
Second, we computed travel delays for all simulated trips. As water levels rise, travel speed decreases, resulting in longer travel times (delays). At the same time, as some roads begin to flood, routes might change as individuals choose alternative, faster routes. Each trip's travel delay is computed given \(t_j-t_j'\), with \(t_j'\) being the base travel time when no rain (and therefore no flooding) occurs. Finally, we converted each trip's travel delays into economic losses using the Danish value of time for travel delays \citep{transportministeriet2022enhedspriser}. Similar to before, economic costs are aggregated as \(D_{i}\) for each TAZ.

\subsubsection{Trip cancellation impacts}
Third, our last type of impact arises when roads or paths become too flooded to allow people to move through, i.e., when there is no possible traversable path between a trip's origin and destination due to water levels being too high. In such cases, we considered affected trips not to occur and treated them as abandoned/cancelled. Assigning a cost to such effects is not easy, but the literature typically assigns 50--100\% of the original cost of making the trip. In our case, we chose to value cancelled trips as 80\% of their original (no-flooded scenario) cost. Importantly, these costs do not reflect the economic cost of the lost activity, but rather reflect an estimate of the value of the trip to the traveler (i.e., individuals would prefer to be travelling and as such there is a net negative value caused by not being able to make the desired trip) \citep{hallenbeckTravelCostsAssociated}. After estimating which trips would be cancelled, we estimated their monetary losses (using a similar value of time to the travel delays case). Finally, we aggregated the costs of cancelled trips at the TAZ level as \(C_i\).

\subsection{Reinforcement Learning}
 
In reinforcement learning, an agent learns a policy for sequential decision-making through repeated interaction with an environment. At each timestep, the agent chooses an intervention, and the IAM simulates the resulting state transition. Specifically, we model this sequential decision problem as a Markov Decision Process (MDP): $\langle \mathcal{S},\mathcal{A},\mathcal{P},r,\gamma\rangle$.
At each decision timestep $t\in\{1,\dots,T\}$, the agent observes a state $s_t\in\mathcal{S}$, samples an action $a_t\in\mathcal{A}$ from a policy $\pi(a\mid s)$, transitions to a new state $s_{t+1}\sim\mathcal{P}(\cdot\mid s_t,a_t)$, and receives a reward $r_t=r(s_t,a_t)$. The objective is to learn a policy that maximizes the expected discounted cumulative reward: 
\begin{equation}
\max_{\pi}\;\mathbb{E}_{\pi}\!\left[\sum_{t=1}^{T}\gamma^{t-1}r_t\right],
\end{equation}
where $\gamma\in(0,1]$ is a discount factor. 

We instantiate the MDP for the climate adaptation problem with the following components: 
\paragraph{\textbf{State space $\mathcal{S}$}} Because flood and transport disruptions exhibit strong spatial dependency, we represent the spatial network as a graph $G=(V,E)$. Nodes $i\in V$ correspond to traffic assignment zones, and edges $(i,j)\in E$ encode spatial proximity. The state consists of node feature vectors $\mathbf{x}_{i,t}$ describing zone $i$ at time $t$. Concretely, we include the three impact components computed by our IAM: direct infrastructure damage $I_{i,t}$, travel delays $D_{i,t}$, and trip cancellations $C_{i,t}$, together with an intervention status vector $\mathbf{z}_{i,t}$ that tracks the remaining effects of implemented interventions (i.e., the volumetric reduction in water volume, $m^3$, decayed over time after deployment). 
\begin{equation}
\mathbf{x}_{i,t}=\big[I_{i,t},\,D_{i,t},\,C_{i,t},\,\mathbf{z}_{i,t}\big].
\end{equation}

\paragraph{\textbf{Action space $\mathcal{A}$}}
Actions are implemented at the zone level. At each timestep $t$, the agent selects, for each zone $i\in V$, one intervention from a discrete set
$a_{i,t}\in\left\{
\begin{aligned}
&\texttt{DoNothing},\ \texttt{Bioretention Planters},\\
&\texttt{Soakaway},\ \texttt{Storage Tank},\ \\ 
&\texttt{Porous Asphalt},\ \texttt{Pervious Concrete}, \ \\
&\texttt{Permeable Pavers},\ \texttt{Grid Pavers} 
\end{aligned}
\right\}.$
yielding the joint action $a_t=\{a_{i,t}\}_{i\in V}$. Once an intervention is deployed in a zone, it remains active for its defined lifetime, and its effectiveness decays over time. To enforce feasibility, interventions currently active in zone $i$ are rendered unavailable via action masking, i.e., the policy is restricted to the subset of actions not implemented in that zone. Each action has distinct properties, including effects, lifetimes, and implementation and maintenance costs. Properties of each action are detailed in \ref{sec_app:rl_actions}.

\paragraph{\textbf{Reward $r$}}
We define the reward as the negative of the total economic cost in each time step: 
\begin{equation}
    R = - \sum_{i \in \texttt{TAZ}} 
    I_{i} +
    D_{i} +
    C_{i} +
    A_{i} +
    M_{i}
\end{equation}
where $I_{i}$, $D_{i}$, and $C_{i}$ are as defined earlier, $A_{i}$ is the aggregated cost of applied intervention across all zones, and $M_{i}$ is the aggregated maintenance cost for all currently active actions.

\paragraph{\textbf{Graph Reinforcement Learning}} To capture spatial correlations in both flood-related impacts and interventions, we parameterise policy $\pi$ with a graph convolutional neural network \cite{kipf2017semisupervised}. Conceptually, the policy implements a graph-to-graph mapping: given the current graph representation of the city, it outputs a node-wise action distribution. Graph neural networks are well-suited to our setting for four main reasons. First, they are permutation-invariant, i.e., the policy's output does not depend on the arbitrary ordering of zones in the input, consistent with the fact that regions in a transportation network have no natural ordering. Second, they learn local operators, allowing the same policy to be applied across graphs of different sizes and topologies. Third, by relying on shared parameters, the policy can be trained and evaluated efficiently as the number of zones increases, enabling scalability to real-world urban transportation networks.

\section{Methodological Validation on a Reduced Problem}
\label{sec:methods_validation}

Climate change adaptation for pluvial flooding is inherently a sequential decision-making problem under deep uncertainty, where investments made today influence economic losses long into the future. The high dimensionality of such problems, arising from stochastic climate events, non-linear system dynamics, and a large space of available adaptation measures, renders classical dynamic programming and static optimization approaches computationally infeasible. Reinforcement learning offers a suitable framework for addressing this challenge, enabling the learning of adaptive, state-dependent policies that optimize long-term cumulative rewards without requiring explicit knowledge of transition probabilities. To showcase our framework relative to other sequence-based optimization approaches, we conduct a set of controlled experiments on reduced problem instances, where direct comparisons with established baselines remain computationally feasible.

\subsection{Bayesian Optimization as a baseline}

Bayesian Optimization (BO) is a family of methods used for global optimization of static decision problems and "black-box" functions (i.e., functions that are expensive to evaluate and lack a known analytical form). In many applications, such function evaluations can be slow or costly, requiring the number of evaluations to be kept small \citep{frazier2018tutorial, xu2025standard}. BO has emerged as a practical tool for optimizing expensive simulation-based models in climate science and energy systems, and as a natural benchmark for evaluating RL algorithms in sequential decision problems \citep{hellan2026bayesian}. However, for high-dimensional policy parameterizations, standard BO (and even improved variants) becomes computationally intractable due to the curse of dimensionality. We therefore compare BO and our RL-based approach on reduced problem instances where BO remains tractable. Importantly, this experiment is not intended as a performance benchmark; rather, it serves as a methodological comparison to examine the implications of adaptivity and sequential decision-making in a controlled setting.

\subsection{Reduced problem definition}

To compare our approach against BO, we define two reduced problem instances in which BO can feasibly identify optimal sets of policies (adaptation measures). \textit{Experiment A} seeks the optimal policy set minimizing impacts and action costs over 5 years across 10 zones, while \textit{Experiment B} restricts the search space to 10 years across 29 zones. Both cases represent scaled-down versions of the whole problem formulated in Section~\ref{sec:methods}, which aims to identify the optimal policy portfolio minimizing impacts and action costs over a 77-year period across 29 zones. 

In the BO algorithm, we used a Gaussian process as a surrogate model with a Matérn 5/2 kernel with automatic relevance determination (ARD). The kernel choice provides a good balance between smoothness and flexibility, making it a robust default choice for this problem \citep{frazier2018tutorial, balandat2020botorch}. As an initial step, 1000 random feasible action plans are generated and evaluated for surrogate model training. For the acquisition function, we used batch Log Expected Improvement (qLogEI), which uses a batch acquisition strategy of Expected Improvement \citep{mockus1975bayesian}, one of the most widely used acquisition functions in BO. For the acquisition function optimization, random, multi-start initialization is used, together with gradient-based refinement. As for stopping criteria, we consider convergence to be achieved when the relative improvement in reward shows no marginal improvement over a window of 100 iterations. Implementation of the BO approach used \texttt{botorch} \citep{balandat2020botorch} and \texttt{gpytorch} \citep{gardner2018gpytorch}. 

As for our RL-based approach, we use the previously defined environment (Section~\ref{sec:methods}) but limit the action space to the first 5/10 years and 10/29 regions. PPO \citep{huang2020closer, schulman2017proximal} was used as the learning algorithm. Additional implementation details are available in Section~\ref{sec:results_implementation}.

\subsection{Results}

Table~\ref{tab:bo_vs_rl} reports the best total impact (sum of flood damages and action costs) achieved by BO and our RL-based approach across three climate scenarios for \textit{Experiment A} (5 years, 10 zones) and \textit{Experiment B} (10 years, 29 zones), respectively.

\begin{table}
    \centering
    \caption{Total reward comparing the BO vs. our RL-based strategy. Average results over three runs are reported ± standard deviations for the two experiments (Exp. A and B) and the three climate scenarios (RCP2.6, 4.5, and 8.5).}
    \label{tab:bo_vs_rl}
    \vspace{3pt}
    \begin{tabular}{llll}
        \toprule
        Exp. & \makecell{Climate\\Scenario} & BO & RL \\
        \midrule
        \midrule
        \multirow[t]{3}{*}{\makecell{A}}
                     & RCP2.6 & -119.58 ± 0.25 
                     & -118.84 ± 1.10 \\
                     & RCP4.5 & -120.10 ± 0.38 
                     & -119.01 ± 0.85 \\
                     & RCP8.5 & -121.24 ± 0.30 
                     & -119.23 ± 1.28 \\[.2cm]
        \multirow[t]{3}{*}{\makecell{B}} 
                     & RCP2.6 & -121.19 ± 1.45 
                     & -117.87 ± 0.57 \\
                     & RCP4.5 & -120.51 ± 1.71 
                     & -117.09 ± 0.64 \\
                     & RCP8.5 & -123.71 ± 1.42
                     & -119.88 ± 1.54 \\
        \bottomrule
    \end{tabular}
\end{table}

Across both case studies and all climate scenarios, our RL-based approach consistently achieves lower total impacts than the BO configuration. In \textit{Experiment A}, RL outperforms BO by approximately $0.6\%$ to $1.7\%$ (73 million (M) DKK to 200 M DKK) depending on the scenario. This advantage is more pronounced in \textit{Experiment B}, where the extended time horizon and larger zone coverage increase problem complexity. Again, the RL approach consistently outperforms BO across all three climate scenarios, with improvements ranging from approximately $2.7\%$ to $3.1\%$ (332 M DKK to 383 M DKK). The widening performance gap between \textit{Experiment A} and \textit{B} suggests that the benefit of RL's adaptive, state-dependent policies grows with problem dimensionality. 

\subsection{Implications for scalability}

The results from \textit{Experiment A} and \textit{Experiment B} support three complementary conclusions. First, RL identifies better-performing policy portfolios than competing baselines, even in reduced settings where such approaches remain tractable. 

Second, the relative advantage of RL scales with problem complexity, reinforcing its suitability for the problem formulated in Section~\ref{sec:methods}. In such a setting, at each step, one single action can be chosen for each of the 29 TAZs. Since each TAZ has 8 possible actions, the number of possible action combinations in a single year is $8^{29}$. Over the whole 77-year horizon, the total number of distinct action plans becomes $(8^{29})^{77} \approx 4 \times 10^{2016}$. Brute force search or traditional dynamic programming approaches quickly become intractable for this problem. This is a typical case of the "curse of dimensionality", where high-dimensional policy and state spaces overwhelm both computational and sample resources.

Third, and perhaps most importantly, the superior performance of RL reflects the value of an adaptive sequential decision-making: RL learns state-dependent strategies that can respond to evolving climate dynamics, leading to lower long-term impacts and costs systematically. Having established the need for adaptive policies in long-term climate adaptation, we now evaluate the proposed RL approach on our full-scale case study.

\section{Full-Scale Case Study: Copenhagen's Inner City}
\label{sec:results}

We now present the main results of the proposed reinforcement learning framework applied to our full-scale case study: Copenhagen's Inner City. We first describe the experimental design, including the simulation setup and evaluation metrics. We then report the performance of the learned adaptation policies in terms of cumulative economic losses and investment patterns, followed by an analysis of the adaptation pathways learned by the RL agent. Finally, we examine how the learnt pathways vary across climate scenarios, highlighting their implications for long-term, adaptive flood risk management.

\subsection{Experimental setup}
\label{sec:results_implementation} 

We set up our IAM using Python, Gymnasium interface \citep{towers_gymnasium_2023}, and Stable-Baseline3 \citep{stable_baselines3}. As previously mentioned, we set our case study to Copenhagen's inner city, consisting of 29 zones where adaptation measures (actions) can be applied at each timestep, and we set the time horizon between 2024 and 2100. For the RL learning algorithm, we use PPO \citep{huang2020closer, schulman2017proximal} with a batch size of 64, 1024 steps per update, 10 epochs for optimizing the surrogate loss, 0.01 as loss entropy coefficient, and a Kullback–Leibler divergence limit of 0.2 to limit updates. Training was performed using 10 parallel environments for a maximum of 4.5 million steps or until the reward function stopped improving (which it did before reaching the maximum number of steps). For all experiments, we report results from 10 runs with different seeds and present the average and standard deviation unless stated otherwise.

\subsection{Reinforcement learning for climate adaptation planning}

We first benchmark the performance of the learned reinforcement learning (RL) policy against two computationally feasible baseline strategies. The \textit{No Control (NC)} strategy applies no adaptation measures throughout the episode, representing a counterfactual scenario in which climate adaptation is not undertaken. The \textit{Random Control (RND)} strategy selects adaptation actions uniformly at random over time and across zones, serving as a non-strategic reference policy. All strategies are evaluated under the intermediate RCP4.5 climate scenario.

\begin{figure*}[!ht]
    \centering
    \includegraphics[width=.98\textwidth]{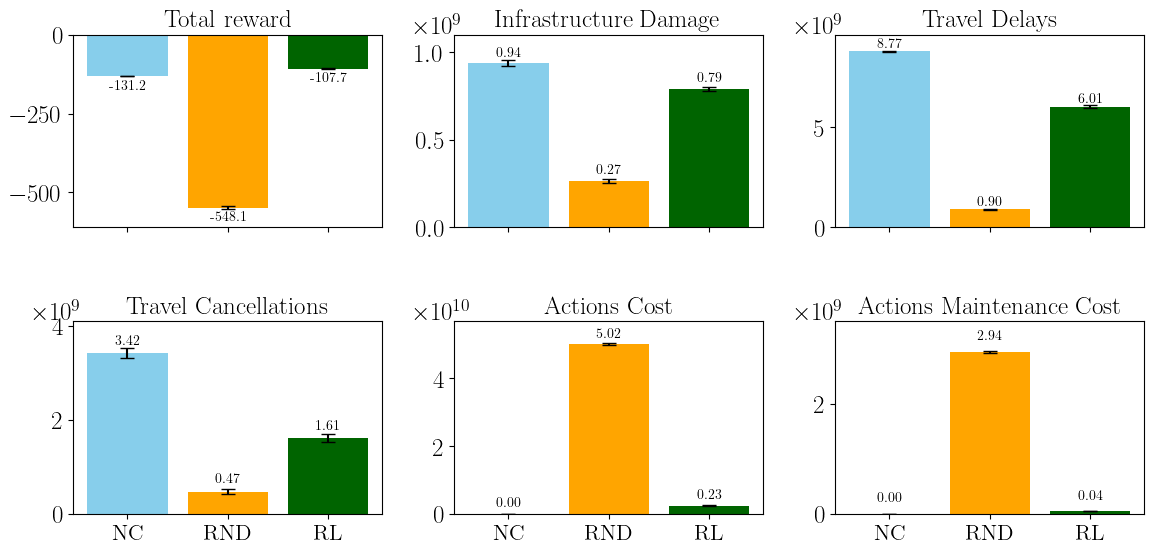} \\
    \caption{Comparison of different adaptation strategies. Total reward (top-left, scaled by \(10^8\)), and reward components are shown: infrastructure damage costs (top-middle), travel delays (top-right), travel cancellations (bottom-left), actions direct costs (bottom-middle), and action maintenance costs (bottom-right). All values correspond to Danish Krone (DKK). Please note the different y-axis scale across figures.}
    \label{fig:benchmark_algorithms}
    \vspace{12pt}
\end{figure*}

Figure~\ref{fig:benchmark_algorithms} reports the average cumulative reward and its decomposition across the five reward components for the three strategies. The RL policy achieves the highest cumulative reward, exceeding the NC strategy by 22\% and the RND strategy by 408\%. While NC and RND constitute intentionally simple baselines, the observed performance gap highlights the importance of coordinated, long-term adaptation decisions. In contrast to the baseline strategies, the RL policy learns to allocate adaptation measures in a temporally and spatially structured strategy, enabling it to balance upfront investment costs against avoided future flood losses effectively. 

This trade-off becomes more noticeable when examining each reward component. The RL strategy reduces infrastructure damage, travel delays, and trip cancellations while incurring moderate adaptation costs. By contrast, the RND strategy achieves larger reductions in individual impact components, but at the expense of substantially higher implementation and maintenance costs, resulting in a lower overall reward. This indicates that non-coordinated adaptation measures fail to balance investment costs against avoided impacts effectively.

Figure~\ref{fig:benchmark_algorithms} shows the temporal evolution of the five reward components over the planning horizon for the three adaptation strategies. The No Control (NC) strategy consistently exhibits high levels of infrastructure damage, travel delays, and trip cancellations, reflecting the absence of any protective measures versus increased impacts over the time horizon. In contrast, both the Random Control (RND) and RL strategies reduce flood-related impacts, although through different adaptation investment strategies.

The RND strategy achieves rapid reductions in impacts during the early timesteps, but this comes at the expense of substantial and persistent action and maintenance costs, driven by uncoordinated and often redundant adaptation measures. As a result, the reduction in impacts is outweighed by the high implementation costs over time. The RL policy, by contrast, exhibits a more gradual and targeted investment pattern, maintaining relatively low action and maintenance costs while achieving reductions in infrastructure damage, travel delays, and travel cancellations. Overall, this shows that the RL agent learns to time and spatially allocate adaptation measures to balance immediate investments with long-term benefits.

\begin{figure*}[!ht]
    \centering
s    \includegraphics[width=.98\textwidth]{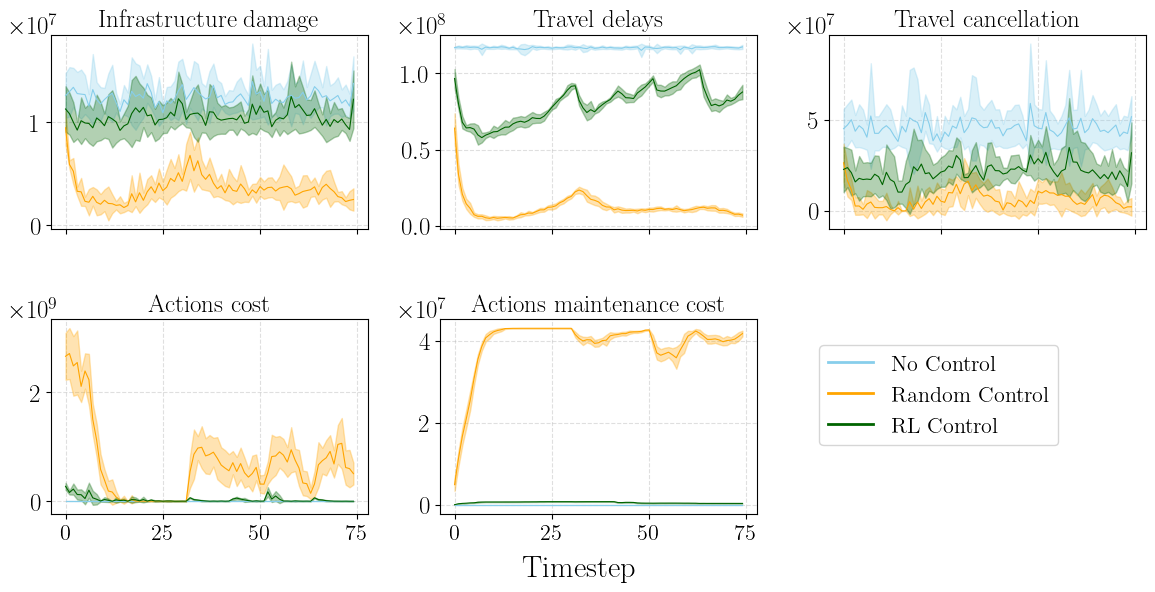} \\
    \caption{Average (and standard deviations shaded) for all five reward components over the period 2024--2100 under the RCP4.5 climate scenario. While Random Control reduces impacts through high and persistent investment costs, the RL policy achieves sustained impact reduction with substantially lower and more stable adaptation expenditures. All values correspond to Danish Krone (DKK). Please note the different y-axis scale across subfigures.}
    \label{fig:benchmark_algorithms_episode}
\end{figure*}

\subsection{Learnt adaptation pathways}

\begin{figure*}[!htb]
    \centering
    \includegraphics[width=.98\textwidth]{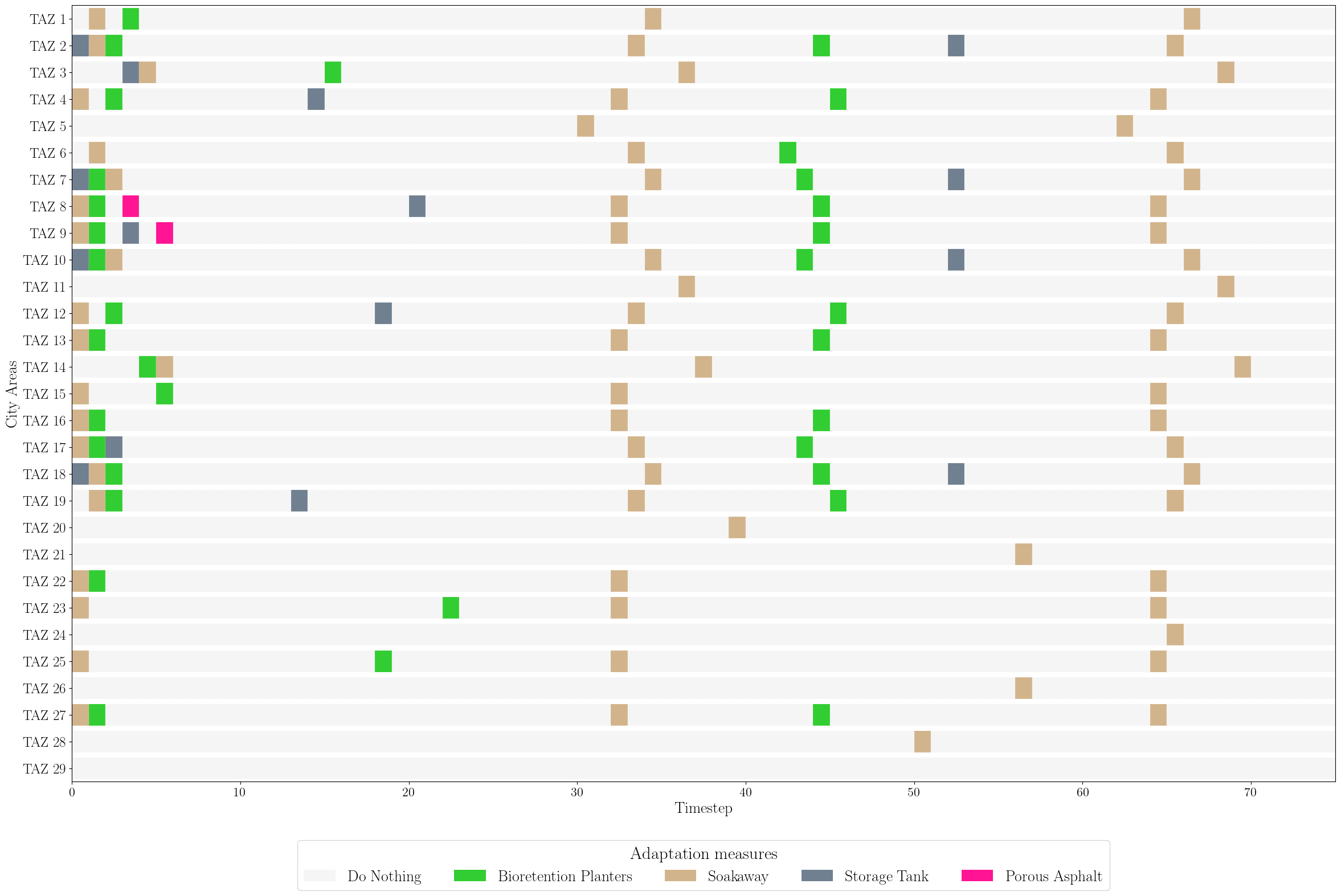} \\
    \caption{Adaptation measures taken over time per zone in Copenhagen's city center for one episode (run) for a RCP4.5 scenario.}
    \label{fig:adaptation_pathway}
\end{figure*}

We now focus on the adaptation pathways learnt by our proposed framework for an RCP4.5 scenario. First, we examine how adaptation measures are implemented across Copenhagen's inner city. Figure~\ref{fig:adaptation_pathway} presents what measures are taken over time for one episode for each one of the zones in Copenhagen's inner city. As we can see, for an RCP4.5 scenario, the RL strategy focuses on using four of the eight possible adaptation measures. \texttt{Soakaways} are the most applied measure (57\% of all actions), followed by \texttt{Bioretention Planters} (28\%), \texttt{Storage Tanks} (13\%), and \texttt{Porous Asphalt} (2\%). On average, 4.41 actions are applied per zone and 1.68 adaptation measures are applied per year. 

\begin{figure*}[!htb]
    \centering
    \includegraphics[width=.98\textwidth]{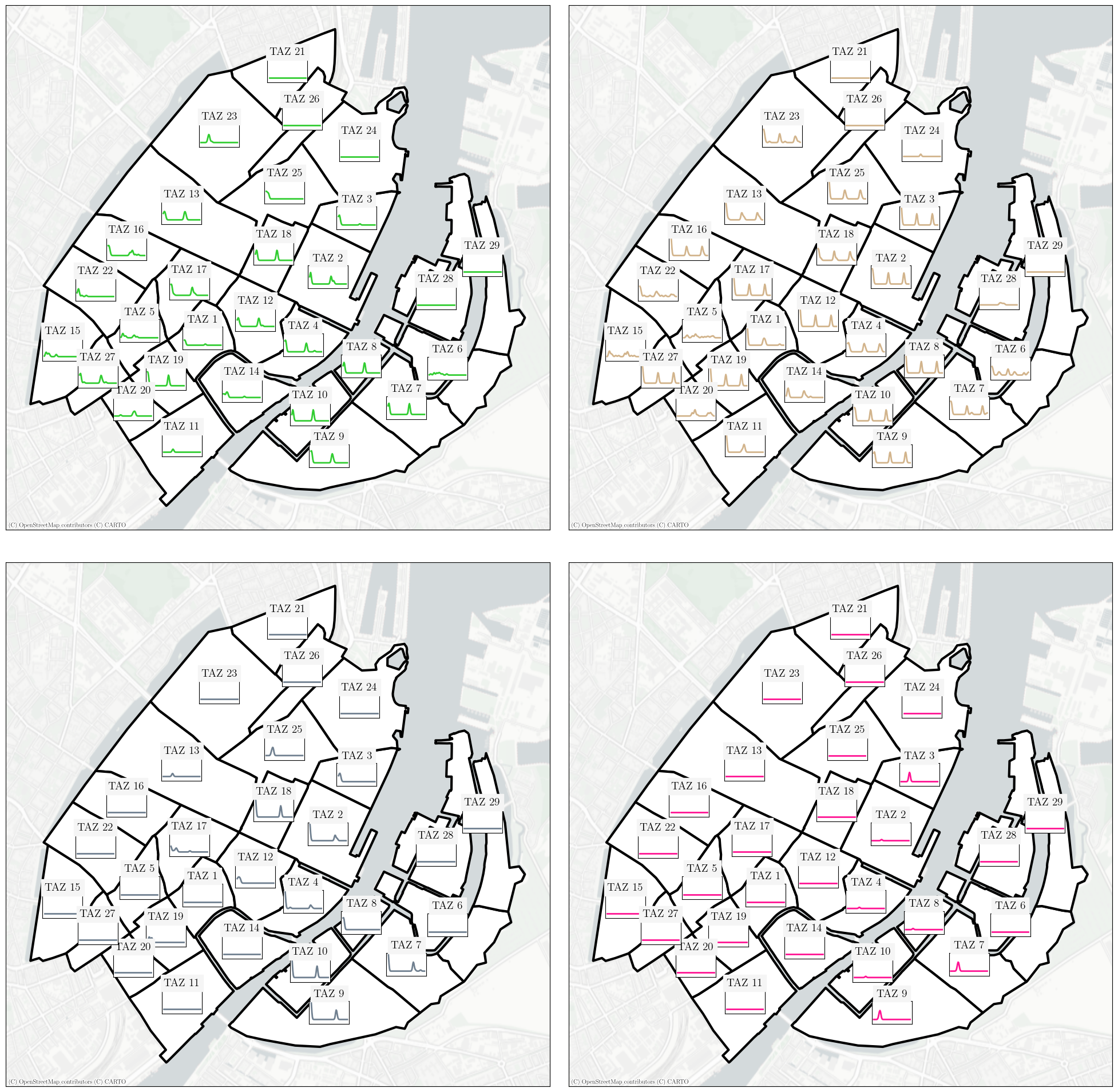} \\
    \caption{Density of actions used over time for 10 runs for a RCP4.5 scenario. \texttt{Bioretention Planters} are shown in top-left, \texttt{Soakaways} in top-right, \texttt{Storage Tanks} in bottom-left, and \texttt{Porous Asphalt} in bottom-right.}
    \label{fig:actions_map}
\end{figure*}

Figure~\ref{fig:actions_map} illustrates the spatial and temporal distribution of four types of adaptation actions across the TAZs under a RCP4.5 scenario, aggregated over 10 simulation runs. Each panel corresponds to one action type and displays, within each TAZ boundary, a time-series plot indicating how frequently and when each action was adopted over the simulation period. Assessing these distribution shifts and geographic contexts also allows for further insights to be captured. First, certain zones exhibit consistently higher densities of specific actions, suggesting a higher need for adaptation measures. Second, while \texttt{Bioretention Planters} and \texttt{Soakaways} are built in nearly all regions in Copenhagen's inner city, \texttt{Storage Tanks} are mainly only used in the central zones, and \texttt{Porous Asphalts} are mainly used in three zones (TAZ 3, 7, and 9) and sometimes in four other zones (TAZ 2, 4, 8, and 10). Third, the temporal profiles show that some actions are deployed in brief, specific periods, whereas others are introduced more gradually over time in response to rainfall events. In sum, the RL strategy dynamically yields different adaptation pathways depending on the current and future expected events.

\subsection{Performance across climate scenarios}

Finally, we analyse how the RL strategy changes across different climate scenarios. First, we examine the total and individual component costs for each RCP scenario: 2.6, 4.5, and 8.5. Table~\ref{tab:results_rcps} enumerates the total costs (i.e., total reward) and component costs (infrastructure damage, travel delays, travel cancellations, action costs, and maintenance costs) for the three climate scenarios. As reported, the overall reward increases in magnitude as the severity of projected warming rises across the three climate scenarios, with RCP8.5 (extreme) showing a 4.6\% increase over RCP4.5 (intermediate) and a 5.6\% increase over RCP2.6 (conservative). Nevertheless, key differences exist between the five cost components. In terms of economic impacts, there is a slight decrease across infrastructure damage, travel delays, and travel cancellations. However, there is a noticeable increase in implemented actions across the three scenarios, with actions in the extreme RCP8.5 scenario accounting for 3.27 billion DKK over 76 years, a 40.9\% increase compared to the RCP4.5 scenario and a 89\% increase compared to the RCP2.6 scenario. This is to be expected, since rainfall intensity increases more severely under the RCP8.5 scenario than under RCP4.5 and 2.6, requiring more measures to mitigate the impacts of more intense rainfall. 

\begin{table*}
    \centering
    \caption{Total reward and costs per reward component using different climate scenarios (RCP2.6, RCP4.5, and RCP8.5) over different 10 runs. Average results are reported ± standard deviations.}
    \label{tab:results_rcps}
    \vspace{3pt}
    \begin{tabular}{lllll}
        \toprule
        \textbf{} & \textbf{RCP2.6} & \textbf{RCP4.5} & \textbf{RCP8.5} \\
        \midrule\midrule
        Infrastructure damage & 8.00 ± 0.18$\times10^8$ & 7.90 ± 0.13$\times10^8$ & 8.05 ± 0.19$\times10^8$ \\
        Travel delays & 6.55 ± 0.10$\times10^9$ & 6.01 ± 0.06$\times10^9$ & 5.54 ± 0.17$\times10^9$ \\
        Travel cancellations & 1.63 ± 0.10$\times10^9$ & 1.61 ± 0.09$\times10^9$ & 1.49 ± 0.10$\times10^9$ \\
        Action costs & 1.73 ± 0.09$\times10^9$ & 2.32 ± 0.11$\times10^9$ & 3.27 ± 0.34$\times10^9$ \\
        Action maintenance costs & 3.46 ± 0.32$\times10^7$ & 4.23 ± 0.21$\times10^7$ & 1.88 ± 0.57$\times10^8$ \\
        \midrule
        Total reward & 1.07 ± 0.02$\times10^{11}$ & 1.08 ± 0.01$\times10^{11}$ & 1.13 ± 0.02$\times10^{11}$ \\
        \bottomrule
    \end{tabular}
\end{table*}

Next, we use our framework to explore the robustness of adaptation strategies to climate uncertainty. Climate projections are rooted in deep uncertainty, so it is vital to estimate and understand how robust adaptation is across different projections. To this end, we perform a set of experiments in which adaptation strategies are designed based on a particular climate scenario belief, but "real/realised" climate projections differ from those anticipated in a counterfactual experiment. In other words, we use our framework to train an RL strategy that follows climate scenario A (e.g., RCP 2.6), and then test it under a different climate scenario B (e.g., RCP 4.5 or RCP 8.5). Table~\ref{tab:train_one_scenario_test_another} reports the cumulative reward across these experiments.

\begin{table}
    \centering
    \caption{Total reward using an RL strategy that was designed using various climate beliefs and reality scenario pairs. Average results are reported ± standard deviations.}
    \label{tab:train_one_scenario_test_another}
    \vspace{3pt}
    \begin{tabular}{llr}
    \toprule
    Belief Scenario & Reality Scenario & Reward \\
    \midrule
    \midrule
    RCP2.6 & RCP2.6 & -107.42 ± 1.55 \\
           & RCP4.5 & -107.53 ± 1.54 \\
           & RCP8.5 & -109.94 ± 1.19 \\
    RCP4.5 & RCP2.6 & -107.42 ± 1.55 \\
           & RCP4.5 & -107.67 ± 1.29 \\
           & RCP8.5 & -109.45 ± 1.15 \\
    RCP8.5 & RCP2.6 & -110.18 ± 1.27 \\
           & RCP4.5 & -110.66 ± 0.90 \\
           & RCP8.5 & -113.03 ± 1.78 \\
    \bottomrule
\end{tabular}

\end{table}

As expected across all belief scenarios, policies achieve their highest rewards when the realized scenario is the mildest (i.e., RCP2.6) and performance degrades when policies trained under milder climate assumptions (e.g., RCP2.6 or RCP4.5) are evaluated under more severe realized scenarios (RCP8.5), reflecting under-adaptation to extreme flooding conditions. Interestingly, strategies following a policy trained for the intermediate scenario (RCP4.5) show the highest average reward across the three scenarios when compared to other policies trained on other scenarios, indicating that such strategy training yields the most cost-effective adaptation strategies. Conversely, policies trained under more severe climate beliefs exhibit comparatively smaller performance losses when evaluated under milder realized scenarios, suggesting a degree of robustness at the expense of higher precautionary investment costs.

Overall, our results highlight a trade-off between performance and robustness: pessimistic climate beliefs lead to more resilient adaptation strategies, while optimistic beliefs yield higher rewards if future conditions remain mild. This underscores the importance of explicitly accounting for climate uncertainty in long-term adaptation planning. These findings also suggest that reinforcement learning provides a flexible framework for exploring adaptive strategies under deep climate uncertainty, while enabling explicit analysis of the consequences of incorrect climate assumptions.

\section{Discussion}
\label{sec:discussion}

\subsection{Methodological implications}

From a methodological perspective, our proposed framework’s findings support our hypothesis that reinforcement learning can be used for climate adaptation planning and add significant value to the exploitation of trade-offs between adaptation measures and climate change-related impacts. From our experiments, we have shown that RL can be used for long-term infrastructure planning problems characterized by delayed rewards, high-dimensional state spaces, stochastic dynamics, and deep uncertainty. Unlike static optimization or rule-based approaches, RL overcomes the limitations of other sequence-based optimization approaches, where uncertainty and high dimensionality (both in time and space) render them ineffective, unfeasible, and untrackable. This framework fills this gap, as it is particularly well-suited to climate adaptation contexts where long-term and trade-off considerations are inherent and must be included, analysed, and understood.

As a whole, our framework serves as a tool to help decision-makers explore and define robust, resilient solutions to minimise disruption to transportation systems caused by floods. It goes beyond defining and mapping disruption paths to helping understand how potential planning opportunities or adaptation measures might translate into long-term solutions that enable more agile, flexible adaptation. In this sense, the proposed reinforcement learning framework should be viewed not as a prescriptive optimization tool but as an exploratory decision-support system that enables the systematic analysis of long-term trade-offs, adaptation pathways, and robustness under climate uncertainty. By learning state-dependent and temporally adaptive policies, the framework provides insights into how different sequences of interventions can shape long-term outcomes, offering decision-makers a structured way to evaluate flexible, resilient strategies that would be difficult to identify with static planning approaches. 

\subsection{Implications for climate adaptation planning}

According to IPCC WGII, long-term, cross-system adaptation policies that integrate multiple climate risk scenarios increase the feasibility, effectiveness, and transformative potential of climate adaptation \citep{ipccClimateChange20222023}. Our results demonstrate that our proposed reinforcement-learning framework effectively captures the sequential and adaptive nature of such long-term flood adaptation planning. Rather than pursuing aggressive early investments, the learned policies allocate adaptation measures gradually, responding to climate events and transportation-related impacts over time. This behavior reflects an implicit balancing of short-term investment costs against long-term risk reduction.

Research is increasingly highlighting the risk of maladaptation to climate change, which is characterised by "locking in" specific vulnerabilities or sub-optimal policies that can be costly or impossible to reverse \citep{ipccClimateChange20222023}. Understanding when, where, and how to invest to make transport resilient to climate change-driven pluvial floods is therefore a vital task for making cities more livable \citep{pregnolato2017impact}. Our findings showcase that there is no "one solution fits all" and adaptation must follow a set of joint measures to produce effective and impactful change. All in all, these trade-offs mirror real-world planning dilemmas and suggest that reinforcement learning provides a valuable framework for exploring robust and adaptive planning strategies under deep uncertainty.

At the same time, it is also crucial to acknowledge the deep uncertainty inherent in climate adaptation planning. While we have shown that our framework can be used across multiple climate scenarios, our experiments have assumed constant scenarios that do not change over time. As cities and countries place greater emphasis (or not) on climate mitigation and adaptation, and as we better understand climate change and improve our climate projection models, our beliefs may change over time and space, suggesting that different adaptation pathways must be pursued. Including such stochastic projections has also enabled more effective pathways to make transport more resilient. 

Nevertheless, there is also a growing need to develop the means to explicitly model the societal trade-offs inherent to adaptation policy \citep{garner_climate_2016}. Adaptation should not focus solely on economic impacts, but also on how different social- and well-being-focused adaptation policies can produce meaningfully different policy pathways compared to more financially focused policies \citep{quinn_health_2023}.

\subsection{Limitations}

This study is not without its limitations. First, our analysis relies on a simulation-based environment that incorporates modeling assumptions for flood dynamics, infrastructure performance, and the mapping of physical impacts to economic losses. While these assumptions are informed by existing literature, models, and data, they shape how the developed environment behaves and, consequently, the learned adaptation policies. As with any simulation-based decision-support tool, the results should therefore be interpreted conditional on the modeled system dynamics rather than direct predictions of real-world outcomes.

Second, the framework explores adaptation strategies under three discrete climate scenarios (RCP2.6, RCP4.5, and RCP8.5). Although this scenario-based approach enables a controlled analysis of climate uncertainty, it does not capture the full range of possible future climate trajectories or their associated probabilities. Incorporating stochastic climate trajectories, scenario ensembles, or probabilistic representations of climate uncertainty would allow for a more comprehensive assessment of robustness. It could lead to adaptation strategies that explicitly account for the likelihood of different climate outcomes.

Third, while our framework overcomes other optimization approaches that are unfeasible in this problem setting, the current implementation is constrained by computational cost. Training reinforcement learning agents over long planning horizons and high-dimensional spatial decision spaces is computationally intensive, limiting the number of adaptation measures, zones (i.e., city areas), and scenarios that can be explored simultaneously. Scaling the framework to larger case studies or a more diverse set of action spaces will require improvements in computational efficiency, for example, through the use of surrogate or metamodels \citep{forrester2008surrogate, razavi2012numerical, razavi2012review, sahlaoui2023review} to approximate expensive simulation components, or through algorithmic advances that reduce training time while preserving policy quality.

Despite these limitations, the framework provides a flexible foundation for exploring adaptive planning strategies, and many of the identified limitations represent opportunities for future methodological extensions rather than fundamental constraints. 

\section{Conclusions}
\label{sec:conclusions}

This study has explored the use of reinforcement learning for long-term climate adaptation. The proposed multi-modular framework enables the learning of adaptive policies that balance investment costs against avoided flood impacts to make transport resilient to the increasing impacts of climate change. Results demonstrate that the learnt policies produce coordinated spatial and temporal adaptation pathways, leading to reductions in infrastructure damage, travel delays, and trip cancellations while maintaining moderate and stable investment costs.

The IPCC has emphasised the importance of developing models that disaggregate climate risk based on RCP or SSP scenarios, that focus on distributional impacts, and that can accommodate inclusive climate governance \citep{ipccClimateChange20222023}. Apart from the leads already discussed in Section~\ref{sec:discussion}, future work should therefore focus on extending the framework to incorporate probabilistic climate information, adaptive belief updating, and multi-objective formulations that explicitly represent social distributional impacts, e.g., on wellbeing or equity. Integrating the approach with stakeholder-driven planning processes also represents a promising direction for enhancing its practical applicability by including local planning contexts or specialized solutions. 

Taken together, the RL framework presented in this paper and our suggested future improvements offer a potentially very valuable avenue to support climate adaptation policy by developing tractable and cross-sectoral adaptation pathways that are robust to multiple climate scenarios.

\section*{Acknowledgements}
\noindent
This work was supported by Villum Fonden grant VIL57387 and Novo Nordisk Foundation grant NNF23OC0085356.

\section*{CRediT (Contributor Roles Taxonomy)}
\noindent
\textbf{Miguel Costa:} Conceptualization, Data curation, Methodology, Formal analysis, Investigation, Software, Writing – original draft.
\textbf{Arthur Vandervoort:} Conceptualization, Data curation, Formal analysis, Investigation, Software, Writing – original draft.
\textbf{Carolin Schmidt:} Methodology, Investigation, Software, Writing – original draft.
\textbf{João Miranda:} Methodology, Software, Writing – original draft.
\textbf{Morten W. Petersen:} Data curation, Methodology, Software.
\textbf{Martin Drews:} Conceptualization, Funding acquisition, Formal analysis, Supervision, Writing – review and editing.
\textbf{Karyn Morrisey:} Conceptualization, Funding acquisition, Formal analysis, Supervision, Writing – review and editing.
\textbf{Francisco C. Pereira:} Conceptualization, Funding acquisition, Formal analysis, Supervision, Writing – review and editing.

\section*{Declaration of generative AI and AI-assisted technologies in the manuscript preparation process.}
\noindent
During the preparation of this work the author(s) used ChatGPT and Claude in order to improve readability and language. After using this tool/service, the author(s) reviewed and edited the content as needed and take(s) full responsibility for the content of the published article.

\bibliographystyle{elsarticle-harv} 
\bibliography{references}

@misc{scalgo,
  author = {SCALGO},
  title = {{SCALGO Live}},
  year  = {2024},
  url   = {https://scalgo.com/live/denmark},
  note = {{Accessed: 2024-06-07}},
}

@misc{danishelevationmodel,
  title = {{Danmarks Højdemodel}},
  author = {{Styrelsen for Dataforsyning og Infrastruktur}},
  year = {2024},
  url = {https://sdfi.dk/data-om-danmark/vores-data/danmarks-hoejdemodel},
  note = {Accessed: 2024-06-07},
}

@article{boeing2024modeling,
  title={Modeling and Analyzing Urban Networks and Amenities with OSMnx},
  author={Boeing, Geoff},
  year={2024},
  publisher={Working paper. URL: \url{https://geoffboeing.com/publications/osmnx-paper/}},
}

@mics{christiansen2021tu,
  title={The Danish National Travel Survey: 0623v1},
  author={Hjalmar Christiansen and Oana Baescu},
  year={2024},
  howpublished = {\url{https://doi.org/10.11581/dtu:00000034}},
  note = {Accessed: 2024-06-01},
}

@mics{vejdirektoratet2022gmm,
  title={Grøn Mobilitetsmodel (GMM)},
  author={Vejdirektoratet},
  year={2022},
  howpublished = {\url{https://www.vejdirektoratet.dk/segment/groen-mobilitetsmodel}},
  note = {Accessed: 2024-06-06},
}

@article{barrington2017world,
  title={The world’s user-generated road map is more than 80\% complete},
  author={Barrington-Leigh, Christopher and Millard-Ball, Adam},
  journal={PloS One},
  volume={12},
  number={8},
  pages={e0180698},
  year={2017},
  publisher={Public Library of Science San Francisco, CA USA}
}

@misc{bicycleaccount2022,
  title = {{The Bicycle Account 2022}},
  author = {{City of Copenhagen}},
  year = {2022},
  url = {https://kk.sites.itera.dk/apps/kk_pub2/pdf/2420_d4db2492337f.pdf},
  note = {},
}

@article{finnis2008field,
  title={Field observations to determine the influence of population size, location and individual factors on pedestrian walking speeds},
  author={Finnis, Kirsten K and Walton, Darren},
  journal={Ergonomics},
  volume={51},
  number={6},
  pages={827--842},
  year={2008},
  publisher={Taylor \& Francis}
}

@TECHREPORT{hallenbeckTravelCostsAssociated,
  AUTHOR =        {Mark E. Hallenbeck and Anne Goodchild and Jerome Drescher},
  TITLE =         "{Travel Costs Associated with Flood Closures of State Highways near Centralia/Chehalis}",
  NUMBER =        {WA-RD 832.1},
  INSTITUTION =   {Washington State Transportation Center},
  MONTH =         {September},
  YEAR  =         {2014},
}

@article{pregnolato2017impact,
    title = {The impact of flooding on road transport: A depth-disruption function},
    journal = {Transportation Research Part D: Transport and Environment},
    volume = {55},
    pages = {67-81},
    year = {2017},
    issn = {1361-9209},
    doi = {https://doi.org/10.1016/j.trd.2017.06.020},
    url = {https://www.sciencedirect.com/science/article/pii/S1361920916308367},
    author = {Maria Pregnolato and Alistair Ford and Sean M. Wilkinson and Richard J. Dawson},
}

@misc{transportministeriet2022enhedspriser,
  title={Transportøkonomiske Enhedspriser, Version 2.0},
  author={Transportministeriet},
  year={2022},
  howpublished = {\url{https://www.man.dtu.dk/myndighedsbetjening/teresa-og-transportoekonomiske-enhedspriser}},
}

@article{ginkel2021flood,
AUTHOR = {van Ginkel, K. C. H. and Dottori, F. and Alfieri, L. and Feyen, L. and Koks, E. E.},
TITLE = {Flood risk assessment of the European road network},
JOURNAL = {Natural Hazards and Earth System Sciences},
VOLUME = {21},
YEAR = {2021},
NUMBER = {3},
PAGES = {1011--1027},
URL = {https://nhess.copernicus.org/articles/21/1011/2021/},
DOI = {10.5194/nhess-21-1011-2021}
}

@book{wallemacq2018economic,
  title={Economic losses, poverty \& disasters: 1998-2017},
  author={Wallemacq, Pascaline and Below, Regina and McClean, Denis},
  year={2018},
  publisher={United Nations Office for Disaster Risk Reduction}
}

@incollection{ipcc2023climate,
  title={Section 3: Long-Term Climate and Development Futures},
  booktitle={Climate Change 2023: Synthesis Report. Contribution of Working Groups I, II and III to the Sixth Assessment Report of the Intergovernmental Panel on Climate Change [Core Writing Team, H. Lee and J. Romero (eds.)]},
  author={{IPCC}},
  year={2023},
  pages={35-115},
  publisher={IPCC, Geneva, Switzerland, doi: 10.59327/IPCC/AR6-9789291691647},
}

@mics{dmi2011adaptation,
  title={Adaptation to the future climate in Denmark},
  author={{Danmarks Meteorologiske Institut}},
  year={2011},
  url = {https://en.klimatilpasning.dk/media/7863/klimatilpasningshæfte%20uk%20web.pdf},
  note = {Accessed: 2024-06-14},
}

@article{yin2016evaluating,
  title={{Evaluating the impact and risk of pluvial flash flood on intra-urban road network: A case study in the city center of Shanghai, China}},
  author={Yin, Jie and Yu, Dapeng and Yin, Zhane and Liu, Min and He, Qing},
  journal={Journal of Hydrology},
  volume={537},
  pages={138--145},
  year={2016},
  publisher={Elsevier}
}

@article{he2021flood,
    title = {{Flood impacts on urban transit and accessibility---A case study of Kinshasa}},
    journal = {Transportation Research Part D: Transport and Environment},
    volume = {96},
    pages = {102889},
    year = {2021},
    issn = {1361-9209},
    doi = {https://doi.org/10.1016/j.trd.2021.102889},
    url = {https://www.sciencedirect.com/science/article/pii/S1361920921001905},
    author = {Yiyi He and Stephan Thies and Paolo Avner and Jun Rentschler},
}

@article{hvass2011sadan,
 year = {2011},
 author  = {Anders Hvass and Mathias Ørsborg Johansen},
 date    = {2011-07-04},
 title   = {Sådan er trafikken ramt af skybruddet},
 journal = {Berlingske},
 url     = {https://www.berlingske.dk/samfund/saadan-er-trafikken-ramt-af-skybruddet},
 note = {Accessed online: 2024-06-06},
}

@article{pregnolato2017climate,
  title={Impact of climate change on disruption to urban transport networks from pluvial flooding},
  author={Pregnolato, Maria and Ford, Alistair and Glenis, Vassilis and Wilkinson, Sean and Dawson, Richard},
  journal={Journal of Infrastructure Systems},
  volume={23},
  number={4},
  pages={04017015},
  year={2017},
  publisher={American Society of Civil Engineers}
}

@article{singh2018vulnerability,
  title={Vulnerability assessment of urban road network from urban flood},
  author={Singh, Prasoon and Sinha, Vinay Shankar Prasad and Vijhani, Ayushi and Pahuja, Neha},
  journal={International Journal of Disaster Risk Reduction},
  volume={28},
  pages={237--250},
  year={2018},
  publisher={Elsevier}
}

@article{wang2020climate,
  title={{Climate change research on transportation systems: Climate risks, adaptation and planning}},
  author={Wang, Tianni and Qu, Zhuohua and Yang, Zaili and Nichol, Timothy and Clarke, Geoff and Ge, Ying-En},
  journal={Transportation Research Part D: Transport and Environment},
  volume={88},
  pages={102553},
  year={2020},
  publisher={Elsevier}
}

@article{gerdes2012what,
 year = {2012},
 author  = {Justin Gerdes},
 date    = {2012-10-31},
 title   = {{What Copenhagen Can Teach Cities About Adapting To Climate Change}},
 journal = {Forbes},
 url     = {https://www.forbes.com/sites/justingerdes/2012/10/31/what-copenhagen-can-teach-cities-about-adapting-to-climate-change/?sh=6c6442961e89},
 note = {Accessed online: 2024-06-06},
}

@article{pregnolato2016assessing,
  title={Assessing urban strategies for reducing the impacts of extreme weather on infrastructure networks},
  author={Pregnolato, Maria and Ford, Alistair and Robson, Craig and Glenis, Vassilis and Barr, Stuart and Dawson, Richard},
  journal={Royal Society open science},
  volume={3},
  number={5},
  pages={160023},
  year={2016},
  publisher={The Royal Society}
}

@article{Hutter01082007,
author = {Gérard Hutter},
title = {{Strategic Planning for Long-Term Flood Risk Management: Some Suggestions for Learning How to Make Strategy at Regional and Local Level}},
journal = {International Planning Studies},
volume = {12},
number = {3},
pages = {273--289},
year = {2007},
publisher = {RSA Website},
doi = {10.1080/13563470701640168},
URL = {https://doi.org/10.1080/13563470701640168},
eprint = {https://doi.org/10.1080/13563470701640168}
}

@book{smith1998floods,
  title={Floods: physical processes and human impacts.},
  author={Smith, Keith and Ward, Roy},
  year={1998},
  publisher="New York: John Wiley"
}

@article{diakakis2020proposal,
title = {Proposal of a flash flood impact severity scale for the classification and mapping of flash flood impacts},
journal = {Journal of Hydrology},
volume = {590},
pages = {125452},
year = {2020},
issn = {0022-1694},
doi = {https://doi.org/10.1016/j.jhydrol.2020.125452},
url = {https://www.sciencedirect.com/science/article/pii/S0022169420309124},
author = {M. Diakakis and G. Deligiannakis and Z. Antoniadis and M. Melaki and N.K. Katsetsiadou and E. Andreadakis and N.I. Spyrou and M. Gogou},
}

@article{georgi2012urban,
  title={{Urban adaptation to climate change in Europe: Challenges and opportunities for cities together with supportive national and European policies}},
  author={Georgi, Birgit and Swart, Rob and Marinova, Natasha and Hove, Bert van and Jacobs, Cor and Klostermann, Judith and Kazmierczak, Aleksandra and Peltonen, Lasse and Kopperoinen, Leena and Oinonen, Kari and others},
  year={2012},
  journal={European Environment Agency}
}

@article{borowska2024changes,
    title = {Changes in intra-city transport accessibility accompanying the occurrence of an urban flood},
    journal = {Transportation Research Part D: Transport and Environment},
    volume = {126},
    pages = {104040},
    year = {2024},
    issn = {1361-9209},
    doi = {https://doi.org/10.1016/j.trd.2023.104040},
    url = {https://www.sciencedirect.com/science/article/pii/S1361920923004376},
    author = {Marta Borowska-Stefańska and Adam Bartnik and Maxim A. Dulebenets and Michał Kowalski and Alireza Sahebgharani and Przemysław Tomalski and Szymon Wiśniewski},
}

@mics{dmi2021van,
  title={Van(d)vittigt mange skybrud over Danmark},
  author={Herdis Preil Damberg},
  year={2021},
  howpublished = {Danmarks Meteorologiske Institut, \url{https://www.dmi.dk/nyheder/2021/vandvittigt-mange-skybrud-over-danmark}},
  note = {Accessed: 2024-06-14},
}

@article{schmith2023regional,
  title={Regional variation of climatological cloudburst frequency estimated from historical observations of daily precipitation sums},
  author={Schmith, Torben and Thejll, Peter and Vejen, Flemming and Christiansen, Bo},
  journal={International Journal of Climatology},
  volume={43},
  number={16},
  pages={7761--7774},
  year={2023},
  publisher={John Wiley \& Sons, Ltd. Chichester, UK}
}

@article{lu2022overview,
title = {{An overview of flood-induced transport disruptions on urban streets and roads in Chinese megacities: Lessons and future agendas}},
journal = {Journal of Environmental Management},
volume = {321},
pages = {115991},
year = {2022},
issn = {0301-4797},
doi = {https://doi.org/10.1016/j.jenvman.2022.115991},
url = {https://www.sciencedirect.com/science/article/pii/S030147972201564X},
author = {Xiaohui Lu and Faith Ka {Shun Chan} and Wei-Qiang Chen and Hing Kai Chan and Xinbing Gu},
}

@techreport{chang2011future,
    author = "Chang, Heejun and Lafrenz, Martin and Jung, Il-Won and Figliozzi, Miguel A and Melgoza, Rolando and Ruelas, David and Platman, Deena and Pederson, Cindy",
    title = "Future flooding impacts on transportation infrastructure and traffic patterns resulting from climate change",
    institution = "OTREC-RR-11-24. Portland, OR: Transportation Research and Education Center (TREC)",
    year = "2010",
    url = "http://dx.doi.org/10.15760/trec.147"
}

@Article{shahdani2022assessing,
    AUTHOR = {Shahdani, Fereshteh Jafari and Santamaria-Ariza, Mónica and Sousa, Hélder S. and Coelho, Mário and Matos, José C.},
    TITLE = {{Assessing Flood Indirect Impacts on Road Transport Networks Applying Mesoscopic Traffic Modelling: The Case Study of Santarém, Portugal}},
    JOURNAL = {Applied Sciences},
    VOLUME = {12},
    YEAR = {2022},
    NUMBER = {6},
    ARTICLE-NUMBER = {3076},
    URL = {https://www.mdpi.com/2076-3417/12/6/3076},
    ISSN = {2076-3417},
    DOI = {10.3390/app12063076}
}

@article{ding_interregional_2023,
	title = {Interregional economic impacts of an extreme storm flood scenario considering transportation interruption: {A} case study of {Shanghai}, {China}},
	volume = {88},
	issn = {22106707},
	shorttitle = {Interregional economic impacts of an extreme storm flood scenario considering transportation interruption},
	url = {https://linkinghub.elsevier.com/retrieve/pii/S221067072200600X},
	doi = {10.1016/j.scs.2022.104296},
	language = {en},
	urldate = {2024-07-04},
	journal = {Sustainable Cities and Society},
	author = {Ding, Wei and Wu, Jidong},
	month = jan,
	year = {2023},
	pages = {104296},
}

@article{markolf2019transportation,
title = {Transportation resilience to climate change and extreme weather events – Beyond risk and robustness},
journal = {Transport Policy},
volume = {74},
pages = {174-186},
year = {2019},
issn = {0967-070X},
doi = {https://doi.org/10.1016/j.tranpol.2018.11.003},
url = {https://www.sciencedirect.com/science/article/pii/S0967070X17305000},
author = {Samuel A. Markolf and Christopher Hoehne and Andrew Fraser and Mikhail V. Chester and B. Shane Underwood},
}

@article{li2024reinforcement,
  title={A reinforcement learning-based routing algorithm for large street networks},
  author={Li, Diya and Zhang, Zhe and Alizadeh, Bahareh and Zhang, Ziyi and Duffield, Nick and Meyer, Michelle A and Thompson, Courtney M and Gao, Huilin and Behzadan, Amir H},
  journal={International Journal of Geographical Information Science},
  volume={38},
  number={2},
  pages={183--215},
  year={2024},
  publisher={Taylor \& Francis}
}

@article{tian2023flooding,
  title={Flooding mitigation through safe \& trustworthy reinforcement learning},
  author={Tian, Wenchong and Xin, Kunlun and Zhang, Zhiyu and Zhao, Muhan and Liao, Zhenliang and Tao, Tao},
  journal={Journal of Hydrology},
  volume={620},
  pages={129435},
  year={2023},
  publisher={Elsevier}
}

@article{bowes2021flood,
  title={Flood mitigation in coastal urban catchments using real-time stormwater infrastructure control and reinforcement learning},
  author={Bowes, Benjamin D and Tavakoli, Arash and Wang, Cheng and Heydarian, Arsalan and Behl, Madhur and Beling, Peter A and Goodall, Jonathan L},
  journal={Journal of Hydroinformatics},
  volume={23},
  number={3},
  pages={529--547},
  year={2021},
  publisher={IWA Publishing}
}

@article{fan2021evaluating,
  title={Evaluating crisis perturbations on urban mobility using adaptive reinforcement learning},
  author={Fan, Chao and Jiang, Xiangqi and Mostafavi, Ali},
  journal={Sustainable Cities and Society},
  volume={75},
  pages={103367},
  year={2021},
  publisher={Elsevier}
}

@inbook{smit2001adaptation,
    title = "Adaptation to climate change in the context of sustainable development and equity",
    author = "B Smit and O Pilifosova and I Burton and B Challenger and S Huq and R Klein and G Yohe and WN Adger and T Downing and E Harvey",
    year = "2001",
    language = "English",
    pages = "877--912",
    editor = "JJ McCarthy and O Canziani and NA Leary and DJ Dokken and KS White",
    booktitle = "Climate Change 2001: Impacts, Adaptation and Vulnerability",
    publisher = "Cambridge University Press",
    address = "United Kingdom",
}

@article{tyler2023ai,
  title={AI tools as science policy advisers? The potential and the pitfalls},
  author={Tyler, Chris and Akerlof, KL and Allegra, Alessandro and Arnold, Zachary and Canino, Henriette and Doornenbal, Marius A and Goldstein, Josh A and Budtz Pedersen, David and Sutherland, William J},
  journal={Nature},
  volume={622},
  number={7981},
  pages={27--30},
  year={2023},
  publisher={Nature Publishing Group UK London}
}

@misc{gilbert2022choicesrisksrewardreports,
      title={Choices, Risks, and Reward Reports: Charting Public Policy for Reinforcement Learning Systems}, 
      author={Thomas Krendl Gilbert and Sarah Dean and Tom Zick and Nathan Lambert},
      year={2022},
      eprint={2202.05716},
      archivePrefix={arXiv},
      primaryClass={cs.LG},
      url={https://arxiv.org/abs/2202.05716}, 
}

@article{matsuo2022deep,
    title = {Deep learning, reinforcement learning, and world models},
    journal = {Neural Networks},
    volume = {152},
    pages = {267-275},
    year = {2022},
    issn = {0893-6080},
    doi = {https://doi.org/10.1016/j.neunet.2022.03.037},
    url = {https://www.sciencedirect.com/science/article/pii/S0893608022001150},
    author = {Yutaka Matsuo and Yann LeCun and Maneesh Sahani and Doina Precup and David Silver and Masashi Sugiyama and Eiji Uchibe and Jun Morimoto},
}

@article{garner_climate_2016,
    title = {Climate risk management requires explicit representation of societal trade-offs},
    volume = {134},
    issn = {1573-1480},
    url = {https://doi.org/10.1007/s10584-016-1607-3},
    doi = {10.1007/s10584-016-1607-3},
    number = {4},
    journal = {Climatic Change},
    author = {Garner, Gregory and Reed, Patrick and Keller, Klaus},
    year = {2016},
    pages = {713--723},
}

@article{cheong2022artificial,
  title={Artificial intelligence for climate change adaptation},
  author={Cheong, So-Min and Sankaran, Kris and Bastani, Hamsa},
  journal={Wiley Interdisciplinary Reviews: Data Mining and Knowledge Discovery},
  volume={12},
  number={5},
  pages={e1459},
  year={2022},
  publisher={Wiley Online Library}
}

@article{rolnick2022tackling,
  title={Tackling climate change with machine learning},
  author={Rolnick, David and Donti, Priya L and Kaack, Lynn H and Kochanski, Kelly and Lacoste, Alexandre and Sankaran, Kris and Ross, Andrew Slavin and Milojevic-Dupont, Nikola and Jaques, Natasha and Waldman-Brown, Anna and others},
  journal={ACM Computing Surveys (CSUR)},
  volume={55},
  number={2},
  pages={1--96},
  year={2022},
  publisher={ACM New York, NY}
}

@article{fend2025reinforcement,
    author = {Kairui Feng  and Ning Lin  and Robert E. Kopp  and Siyuan Xian  and Michael Oppenheimer },
    title = {Reinforcement learning–based adaptive strategies for climate change adaptation: An application for coastal flood risk management},
    journal = {Proceedings of the National Academy of Sciences},
    volume = {122},
    number = {12},
    pages = {e2402826122},
    year = {2025},
    doi = {10.1073/pnas.2402826122},
    URL = {https://www.pnas.org/doi/abs/10.1073/pnas.2402826122},
    eprint = {https://www.pnas.org/doi/pdf/10.1073/pnas.2402826122},
}

@article{GARNER201896,
    title = {Using direct policy search to identify robust strategies in adapting to uncertain sea-level rise and storm surge},
    journal = {Environmental Modelling \& Software},
    volume = {107},
    pages = {96-104},
    year = {2018},
    issn = {1364-8152},
    doi = {https://doi.org/10.1016/j.envsoft.2018.05.006},
    url = {https://www.sciencedirect.com/science/article/pii/S1364815217307727},
    author = {Gregory G. Garner and Klaus Keller},
}

@article{LICKLEY201418,
    title = {Analysis of coastal protection under rising flood risk},
    journal = {Climate Risk Management},
    volume = {6},
    pages = {18-26},
    year = {2014},
    issn = {2212-0963},
    doi = {https://doi.org/10.1016/j.crm.2015.01.001},
    url = {https://www.sciencedirect.com/science/article/pii/S2212096315000029},
    author = {Megan J. Lickley and Ning Lin and Henry D. Jacoby},
}

@article{sobhaniyeh2021robust,
  title={Robust flood risk management strategies through bayesian estimation and multi-objective optimization},
  author={Sobhaniyeh, Zahra and Niksokhan, Mohammad Hossein and Omidvar, Babak and Gaskin, Susan},
  journal={International journal of environmental research},
  volume={15},
  number={6},
  pages={1057--1070},
  year={2021},
  publisher={Springer}
}

@misc{towers_gymnasium_2023,
        title = {Gymnasium},
        url = {https://zenodo.org/record/8127025},
        urldate = {2023-07-08},
        publisher = {Zenodo},
        author = {Towers, Mark and Terry, Jordan K. and Kwiatkowski, Ariel and Balis, John U. and Cola, Gianluca de and Deleu, Tristan and Goulão, Manuel and Kallinteris, Andreas and KG, Arjun and Krimmel, Markus and Perez-Vicente, Rodrigo and Pierré, Andrea and Schulhoff, Sander and Tai, Jun Jet and Shen, Andrew Tan Jin and Younis, Omar G.},
        month = mar,
        year = {2023},
        doi = {10.5281/zenodo.8127026},
}

@article{stable_baselines3,
  author  = {Antonin Raffin and Ashley Hill and Adam Gleave and Anssi Kanervisto and Maximilian Ernestus and Noah Dormann},
  title   = {{Stable-Baselines3: Reliable Reinforcement Learning Implementations}},
  journal = {Journal of Machine Learning Research},
  year    = {2021},
  volume  = {22},
  number  = {268},
  pages   = {1-8},
  url     = {http://jmlr.org/papers/v22/20-1364.html}
}

@article{huang2020closer,
  title={A closer look at invalid action masking in policy gradient algorithms},
  author={Huang, Shengyi and Onta{\~n}{\'o}n, Santiago},
  journal={arXiv preprint arXiv:2006.14171},
  year={2020}
}

@article{schulman2017proximal,
  title={Proximal policy optimization algorithms},
  author={Schulman, John and Wolski, Filip and Dhariwal, Prafulla and Radford, Alec and Klimov, Oleg},
  journal={arXiv preprint arXiv:1707.06347},
  year={2017}
}

@article{quinn_health_2023,
    title = {Health and wellbeing implications of adaptation to flood risk},
    volume = {52},
    issn = {1654-7209},
    url = {https://link.springer.com/article/10.1007/s13280-023-01834-3},
    doi = {10.1007/s13280-023-01834-3},
    number = {5},
    journal = {Ambio},
    author = {Quinn, Tara and Heath, Stacey and Adger, W. Neil and Abu, Mumuni and Butler, Catherine and Codjoe, Samuel Nii Ardey and Horvath, Csaba and Martinez-Juarez, Pablo and Morrissey, Karyn and Murphy, Conor and Smith, Richard},
    year = {2023},
    pages = {952--962},
}

@article{vanvuuren2011,
  title={The representative concentration pathways: an overview},
  author={Van Vuuren, Detlef P and Edmonds, Jae and Kainuma, Mikiko and Riahi, Keywan and Thomson, Allison and Hibbard, Kathy and Hurtt, George C and Kram, Tom and Krey, Volker and Lamarque, Jean-Francois and others},
  journal={Climatic Change},
  volume={109},
  pages={5--31},
  year={2011},
  publisher={Springer}
}

@misc{dmi2023klimaatlas,
  title={Klimaatlas},
  author={{Danmarks Meteorologiske Institut}},
  year={2023},
  url = {https://www.dmi.dk/klima-atlas/data-i-klimaatlas},
  note = {{Accessed: 2024-08-26}},
}

@inproceedings{
kipf2017semisupervised,
title={Semi-Supervised Classification with Graph Convolutional Networks},
author={Thomas N. Kipf and Max Welling},
booktitle={International Conference on Learning Representations},
year={2017},
url={https://openreview.net/forum?id=SJU4ayYgl}
}

@misc{frazier2018tutorial,
      title={A Tutorial on Bayesian Optimization}, 
      author={Peter I. Frazier},
      year={2018},
      eprint={1807.02811},
      archivePrefix={arXiv},
      primaryClass={stat.ML},
      url={https://arxiv.org/abs/1807.02811}, 
}

@misc{xu2025standard,
      title={Standard Gaussian Process is All You Need for High-Dimensional Bayesian Optimization}, 
      author={Zhitong Xu and Haitao Wang and Jeff M Phillips and Shandian Zhe},
      year={2025},
      eprint={2402.02746},
      archivePrefix={arXiv},
      primaryClass={cs.LG},
      url={https://arxiv.org/abs/2402.02746}, 
}

@inproceedings{hellan2026bayesian,
    author="Hellan, Sigrid Passano
    and Lucas, Christopher G.
    and Goddard, Nigel H.",
    editor="Zhang, Yingqian
    and Hladik, Milan
    and Moosaei, Hossein",
    title="Bayesian Optimisation Against Climate Change: Applications and Benchmarks",
    booktitle="Learning and Intelligent Optimization",
    year="2026",
    publisher="Springer Nature Switzerland",
    address="Cham",
    pages="107--118",
    isbn="978-3-032-09156-7"
}

@article{gardner2018gpytorch,
  title={Gpytorch: Blackbox matrix-matrix gaussian process inference with gpu acceleration},
  author={Gardner, Jacob and Pleiss, Geoff and Weinberger, Kilian Q and Bindel, David and Wilson, Andrew G},
  journal={Advances in Neural Information Processing Systems},
  volume={31},
  year={2018}
}

@inproceedings{balandat2020botorch,
  title = {{BoTorch: A Framework for Efficient Monte-Carlo Bayesian Optimization}},
  author = {Balandat, Maximilian and Karrer, Brian and Jiang, Daniel R. and Daulton, Samuel and Letham, Benjamin and Wilson, Andrew Gordon and Bakshy, Eytan},
  booktitle = {Advances in Neural Information Processing Systems 33},
  year = 2020,
  url = {http://arxiv.org/abs/1910.06403}
}

@InProceedings{mockus1975bayesian,
    author="Mo{\v{c}}kus, J.",
    editor="Marchuk, G. I.",
    title="On bayesian methods for seeking the extremum",
    booktitle="Optimization Techniques IFIP Technical Conference Novosibirsk, July 1--7, 1974",
    year="1975",
    publisher="Springer Berlin Heidelberg",
    address="Berlin, Heidelberg",
    pages="400--404",
    isbn="978-3-540-37497-8"
}

@book{forrester2008surrogate,
  title        = {Engineering Design via Surrogate Modelling: A Practical Guide},
  author       = {Forrester, Alexander and Sobester, Andr{\'a}s and Keane, Andy},
  publisher    = {Wiley},
  year         = {2008},
}

@article{razavi2012numerical,
  title={Numerical assessment of metamodelling strategies in computationally intensive optimization},
  author={Razavi, Saman and Tolson, Bryan A and Burn, Donald H},
  journal={Environmental Modelling \& Software},
  volume={34},
  pages={67--86},
  year={2012},
  publisher={Elsevier}
}

@article{razavi2012review,
  title={Review of surrogate modeling in water resources},
  author={Razavi, Saman and Tolson, Bryan A and Burn, Donald H},
  journal={Water Resources Research},
  volume={48},
  number={7},
  year={2012},
  publisher={Wiley Online Library}
}

@article{sahlaoui2023review,
  title={A review on simulation-based metamodeling in emergency healthcare: methodology, applications, and future challenges},
  author={Sahlaoui, Fatima-Zahra and Aboueljinane, Lina and Lebbar, Maria},
  journal={Simulation},
  volume={99},
  number={10},
  pages={989--1009},
  year={2023},
  publisher={SAGE Publications Sage UK: London, England}
}

@incollection{intergovernmentalpanelonclimatechangeipccClimateChange20222023a,
  title = {Annex {{II}}: {{Glossary}}},
  booktitle = {Climate {{Change}} 2022 -- {{Impacts}}, {{Adaptation}} and {{Vulnerability}}: {{Working Group II Contribution}} to the {{Sixth Assessment Report}} of the {{Intergovernmental Panel}} on {{Climate Change}}},
  author = {{IPCC}},
  year = 2023,
  month = jun,
  edition = {1},
  publisher = {Cambridge University Press},
  doi = {10.1017/9781009325844},
  urldate = {2026-03-02},
  copyright = {https://www.cambridge.org/core/terms},
  isbn = {978-1-009-32584-4}
}

@book{ipccClimateChange20222023,
  title = {Climate {{Change}} 2022 -- {{Impacts}}, {{Adaptation}} and {{Vulnerability}}: {{Working Group II Contribution}} to the {{Sixth Assessment Report}} of the {{Intergovernmental Panel}} on {{Climate Change}}},
  shorttitle = {Climate {{Change}} 2022 -- {{Impacts}}, {{Adaptation}} and {{Vulnerability}}},
  author = {{IPCC}},
  year = 2023,
  month = jun,
  edition = {1},
  publisher = {Cambridge University Press},
  doi = {10.1017/9781009325844},
  urldate = {2025-01-24},
  copyright = {https://www.cambridge.org/core/terms},
  isbn = {978-1-009-32584-4},
  langid = {english}
}

@inproceedings{permeable2015permeable,
  title={Permeable pavements},
  author={{Permeable Pavements Task Committee}},
  year={2015},
  organization={American Society of Civil Engineers}
}

@mastersthesis{cuaran2015design,
  author  = {Cuaran Bermudez, Andres Felipe and Lundberg, Linnea},
  title   = {Design of Bioretention Planters for Stormwater Flow control and removal of Toxic Metals and Organic Contaminats.},
  school  = {Department of Energy and Environment, Chalmers University of Technology},
  year    = {2015}
}

@article{stovinRetrofitSuDSCost2007,
	title = {Retrofit {SuDS}—cost estimates and decision-support tools},
	volume = {160},
	issn = {1741-7589, 1751-7729},
	url = {https://www.icevirtuallibrary.com/doi/10.1680/wama.2007.160.4.207},
	doi = {10.1680/wama.2007.160.4.207},
	language = {en},
	number = {4},
	urldate = {2025-06-13},
	journal = {Proceedings of the Institution of Civil Engineers - Water Management},
	author = {Stovin, V. R. and Swan, A. D.},
	month = dec,
	year = {2007},
	pages = {207--214},
}

\appendix

\section{Details of adaptation measures}
\label{sec_app:rl_actions}

We include eight adaptation measures (RL actions) in our framework, which are typically used as climate change adaptation solutions:
\begin{description}
    \item[Do Nothing] Implement no adaptation measures.
    
    \item[Bioretention Planters] Multi-layered permeable systems that capture water runoff and the top layer is covered in some vegetation \citep{cuaran2015design}.
    
    \item[Soakaways] Underground Structure filled with granular material designed to store rapid runoff and allow efficient infiltration into the surrouding soil \citep{stovinRetrofitSuDSCost2007}.
    
    \item[Concrete Storage Tanks] Underground reinforcement concrete tanks to attenuate flow and capture peak flow and runoff \citep{stovinRetrofitSuDSCost2007}.
    
    \item[Porous Asphalt] Permeable asphalt surface with an underlain open-graded aggregate choker course and a reservoir bed. Porous asphalt systems allow for stormwater filtration/infiltration and storage \citep{permeable2015permeable}.
    
    \item[Pervious Concrete] Hydraulic cementitious binding system combined with an open-graded aggregate to produce a durable pavement that allows rapid infiltration of stormwater \citep{permeable2015permeable}. 
    
    \item[Permeable interlocking concrete pavement (PICP)]  Manufactured impervious concrete units that are designed with small permeable joints where water can infiltrate
    \citep{permeable2015permeable}.
    
    \item[Grid Pavers] Concrete or plastic open-celled paving units typically covered in topsoild or grass and where water can infiltrate through “cells” or openings \citep{permeable2015permeable}.
    
\end{description}

\setcellgapes{6pt}
\makegapedcells
\begin{table*}[htb]
\caption{Adaptation measure details used in our framework.}
\label{tab:app_action_details}
\begin{tabular}{p{1.9cm}p{3cm}p{3cm}p{3cm}p{3cm}p{1.6cm}}
\hline
\textbf{Adaptation Measure} & \textbf{Effect}                 & \textbf{Implementation details}         & \textbf{Implementation costs (DKK)} & \textbf{Maintenace costs (DKK/year)} & \textbf{Lifetime (years)} \\ \hline\hline
Bioretention planter        & Captures 2 $m^3$ of water runoff   & Applied every 30m of road               & 14312 per planter                   & 24 per planter                       & 40                        \\
Soakaway                    & Captures 5.4 $m^3$ of water runoff & Applied once per road                   & 7273                                & 1.9                                  & 30                        \\
Storage tank                & Captures 15 $m^3$ of water runoff  & Applied once per road                   & 104896                              & 5                                    & 50                        \\
Porous asphalt              & Captures 0.3m of water depth    & Replaces the road with this pavement type & 995.77 per m2 of surface area       & 0.635 per $m^2$ of surface area         & 30                        \\
Pervious concrete           & Captures 0.45m of water depth   & Replaces the road with this pavement type & 1199.81 per m2 of surface area      & 0.635 per $m^2$ of surface area         & 30                        \\
PICP                        & Captures 0.25m of water depth   & Replaces the road with this pavement type & 1046.78 per m2 of surface area      & 5.195 per $m^2$ of surface area         & 50                        \\
Grid pavers                 & Captures 0.2m of water depth    & Replaces the road with this pavement type & 1097.79 per m2 of surface area      & 5.195 per $m^2$ of surface area         & 30  \\   
\hline
\end{tabular}
\end{table*}

Each adaptation solution is characterized by having the properties in Table~\ref{tab:app_action_details}.



\end{document}